\documentclass[10pt,twocolumn,letterpaper]{article}

 \usepackage[pagenumbers]{iccv} % To force page numbers, e.g. for an arXiv version

\definecolor{iccvblue}{rgb}{0.21,0.49,0.74}
\usepackage[pagebackref,breaklinks,colorlinks,citecolor=iccvblue]{hyperref}
\usepackage{caption}
\usepackage{pifont}
\usepackage{multirow}
\usepackage{colortbl}
\usepackage{tabularx}
\usepackage{array} 
\usepackage{graphicx}
\usepackage[accsupp]{axessibility}

\definecolor{best}{rgb}{1, 0.5, 0}
\definecolor{second}{rgb}{1, 1, 0}

\usepackage[capitalize]{cleveref}
\crefname{section}{Sec.}{Secs.}
\Crefname{section}{Section}{Sections}
\Crefname{table}{Table}{Tables}
\crefname{table}{Tab.}{Tabs.}

\def\paperID{2539} % *** Enter the Paper ID here
\def\confName{ICCV}
\def\confYear{2025}

\title{CoMoGaussian: Continuous Motion-Aware Gaussian Splatting \\from Motion-Blurred Images\vspace{-3mm}}

\author{{Jungho Lee$^1$ \quad Donghyeong Kim$^1$ \quad Dogyoon Lee$^1$ \quad Suhwan Cho$^1$ \quad Minhyeok Lee$^1$\vspace{1mm}}\\ {Wonjoon Lee$^1$ \quad Taeoh Kim$^2$ \quad Dongyoon Wee$^2$ \quad Sangyoun Lee$^1$\vspace{-3mm}}\\ \\ 
	{$^1$Yonsei University} \quad\quad {$^2$NAVER Cloud}\\
}

\begin{document}
\twocolumn[{
	\renewcommand\twocolumn[1][]{#1}
	\maketitle
	\begin{center}
		\centering
		\captionsetup{type=figure}
		\vspace{-5mm}
		\includegraphics[width=1\linewidth]{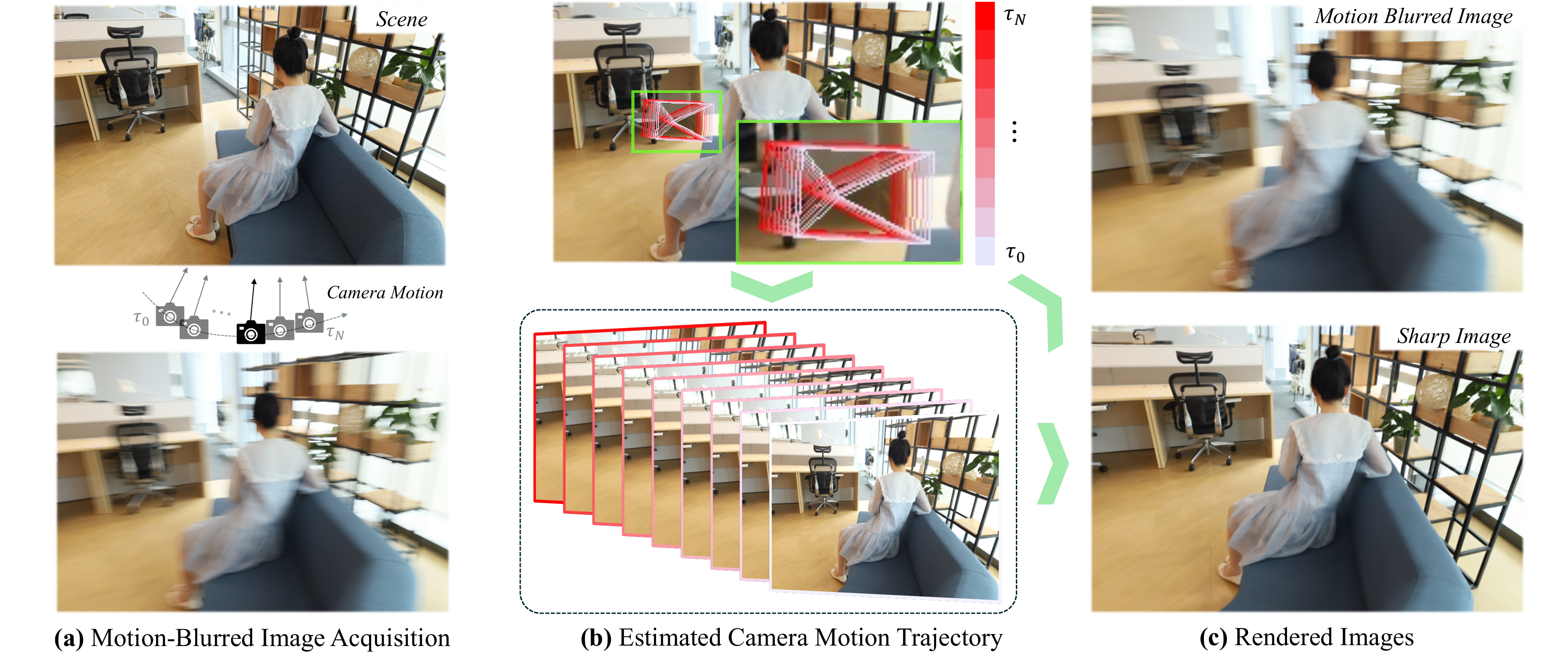}
		\vspace{-6mm}
		\caption{We propose CoMoGaussian, a novel framework for reconstructing 3D scenes from camera motion-blurred images. \textbf{(a)} Camera motion-blurred images are generated by the continuous movement of the camera during the exposure time. \textbf{(b)} The top image represents the camera poses obtained through CoMoGaussian, showing a continuous trajectory indicated by gradual change of colors. The bottom images are the rendering outputs from each camera pose. \textbf{(c)} By aggregating the rendered images obtained in step (b), we get the motion-blurred output image shown at the top, while the sharp output image is rendered from the given calibrated camera pose.}
		\label{fig:1_teaser}
	\end{center}
}]
\begin{abstract}
3D Gaussian Splatting (3DGS) has gained significant attention due to its high-quality novel view rendering, motivating research to address real-world challenges. 
A critical issue is the camera motion blur caused by movement during exposure, which hinders accurate 3D scene reconstruction. 
In this study, we propose CoMoGaussian, a \textbf{Co}ntinuous \textbf{Mo}tion-Aware \textbf{Gaussian} Splatting that reconstructs precise 3D scenes from motion-blurred images while maintaining real-time rendering speed. 
Considering the complex motion patterns inherent in real-world camera movements, we predict continuous camera trajectories using neural ordinary differential equations (ODEs). 
To ensure accurate modeling, we employ rigid body transformations, preserving the shape and size of the object but rely on the discrete integration of sampled frames. 
To better approximate the continuous nature of motion blur, we introduce a continuous motion refinement (CMR) transformation that refines rigid transformations by incorporating additional learnable parameters.
By revisiting fundamental camera theory and leveraging advanced neural ODE techniques, we achieve precise modeling of continuous camera trajectories, leading to improved reconstruction accuracy. 
Extensive experiments demonstrate state-of-the-art performance both quantitatively and qualitatively on benchmark datasets, which include a wide range of motion blur scenarios, from moderate to extreme blur.
Project page is available at {\href{https://Jho-Yonsei.github.io/CoMoGaussian/}{\textcolor{iccvblue}{https://Jho-Yonsei.github.io/CoMoGaussian}}}.
\end{abstract}    
\section{Introduction}
\label{sec:intro}

Novel view synthesis has recently garnered significant attention, with neural radiance fields (NeRF)~\cite{mildenhall2020nerf} making considerable advancements in photo-realistic neural rendering. 
NeRF takes sharp 2D images from multiple views as input to reconstruct precise 3D scenes, which are crucial for applications such as augmented reality (AR) and virtual reality (VR). 
%NeRF models radiance and volume density through an implicit neural representation using multi-layer perceptrons (MLPs), based on the coordinates and viewing directions of 3D points. 
However, inefficient memory usage of NeRF's volume rendering process poses challenges for real-time applications. 
To address this, 3D Gaussian Splatting (3DGS)~\cite{kerbl20233d} has recently emerged, offering an alternative by explicitly representing 3D scenes and enabling real-time rendering through a differentiable splatting method.

However, for precise 3D scene representation in real-world scenarios, it is essential to address various forms of image quality degradation, such as camera motion blur and defocus blur. 
Current methods like NeRF and 3DGS rely on sharp images as input, which assumes highly ideal conditions. 
Obtaining sharp images requires a large depth of field (DoF), necessitating a very small aperture setting~\cite{hecht2012optics}. 
However, a smaller aperture limits light intake, leading to longer exposure times. 
With longer exposure, even slight camera movement introduces complex motion blur in the images. 
Therefore, developing techniques that can handle motion blur in input images is crucial for accurate 3D scene reconstruction. 
Our research tackles this challenge by focusing on reconstructing precise 3D scenes from motion-blurred images, thus broadening the applicability of neural rendering in more realistic conditions.

Recently, several methods have been proposed to achieve sharp novel view rendering from camera motion-blurred images. 
Inspired by traditional blind image deblurring techniques, Deblur-NeRF~\cite{ma2022deblurnerf} firstly introduces a method for deblurring 3D scenes. 
This approach designs a learnable kernel that intentionally blurs images during training, and during rendering, excludes the learned blurring kernel to render sharp novel view images. 
Following this approach, various methods targeting camera motion blur~\cite{lee2023dp,lee2024deblurring,wang2023bad,zhao2024bad,peng2023pdrf,peng2024bags} have emerged, aiming to improve blurring kernel estimation accuracy. 
However, all of these methods predict the camera trajectory without enforcing continuity of camera motion throughout the exposure time (\textit{e.g}., simple spline functions~\cite{wang2023bad,zhao2024bad}). 
This oversight is critical because motion blur arises from the continuous integration of scene radiance over the camera’s actual movement during the exposure interval. 
When continuity is not properly accounted for, the predicted motion can exhibit abrupt or piecewise transitions that deviate from physically plausible camera trajectories, resulting in inaccurate blur approximation and suboptimal deblurring performance. 
While these methods can model simpler motion blur scenarios, they fail to capture complex, smoothly varying camera motion due to their lack of a temporally coherent representation.

In this paper, we propose a \textbf{Co}ntinuous \textbf{Mo}tion-Aware \textbf{Gaussian} Splatting, namly CoMoGaussian, a novel approach with three key contributions. 
First, we apply neural ordinary differential equations (ODEs)~\cite{chen2018neural} to model continuous camera movement during exposure time, as illustrated in~\cref{fig:1_teaser}. 
By continuously modeling the camera trajectory in 3D space, we introduce a model that is fundamentally different from existing methods~\cite{ma2022deblurnerf,lee2023dp,peng2023pdrf,wang2023bad}. 
Second, we continuously model rigid body transformation over time to accurately capture the shape and size of the static subject throughout the camera movement.
By leveraging a continuous representation of rigid motion, our approach better accounts for subtle variations in trajectory, leading to more precise reconstruction of static structures.
Third, we introduce a continuous motion refinement (CMR) transformation, which enhances rigid motion modeling by incorporating learnable transformations, enabling more accurate approximation of motion blur trajectories.
To integrate these components into a high-quality real-time rendering framework, we adopt Mip-Splatting~\cite{yu2024mip}, a differentiable rasterization-based approach built on 3DGS~\cite{kerbl20233d}.
We evaluate and compare our approach on Deblur-NeRF~\cite{ma2022deblurnerf} and ExbluRF~\cite{lee2023exblurf} datasets, achieving state-of-the-art results across the benchmarks. 
To demonstrate the effectiveness of CoMoGaussian, we conduct various ablative experiments on the proposed contributions.

%While neural ODEs allow for continuous modeling of camera motion, the discretization required for numerical integration inevitably introduces approximation errors in reconstructing motion blur.
%For superior rendering quality with real-time rendering speed, we employ Mip-Splatting~\cite{yu2024mip}, a differentiable rasterization-based method based on 3DGS~\cite{kerbl20233d}, rather than ray tracing-based methods (\textit{e.g}., NeRF~\cite{mildenhall2020nerf}, TensoRF~\cite{chen2022tensorf}).
\section{Related Work}
\label{sec:relatedwork}

\subsection{Neural Rendering}
Neural rendering has seen explode in the fields of computer graphics and vision area thanks to the emergence of ray tracing-based neural radiance fields~(NeRF)~\cite{mildenhall2020nerf}, which provide realistic rendering quality from 3D scenes. 
NeRF has led to a wide range of studies, including TensoRF~\cite{chen2022tensorf}, Plenoxel~\cite{fridovich2022plenoxels}, and Plenoctree~\cite{yu2021plenoctrees}, which have aimed at overcoming slow rendering speed.
Among these advancements, the recent emergence of 3D Gaussian splatting~(3DGS)~\cite{kerbl20233d}, which offers remarkable performance along with fast training and rendering speed, has further accelerated research in the neural rendering. 
Additionally, there has been active research focusing on the non-ideal conditions of given images, such as sparse-view images~\cite{niemeyer2022regnerf,yang2023freenerf,wang2023sparsenerf}, and the absence of camera parameters~\cite{wang2021nerf,bian2023nope}.
Moreover, there has been significant research attention on non-ideal conditions inherent to the images themselves, such as low-light~\cite{mildenhall2022nerfinthedark,pearl2022nan}, blur~\cite{ma2022deblurnerf,lee2023dp,peng2023pdrf,wang2023bad,lee2024smurf,zhao2024bad,peng2024bags}. 
Recently, neural rendering from blurry images has attracted attention due to its practical applicability. 

\subsection{Neural Rendering from Blurry Images}
Deblur-NeRF~\cite{ma2022deblurnerf} firstly introduced the deblurred neural radiance fields by importing the blind-deblurring mechanism into the NeRF framework. 
They introduce specific blur kernel in front of the NeRF framework imitating the blind deblurring in 2D image deblurring area.
After the emergence of~\cite{ma2022deblurnerf}, several attempts have been proposed to model the precise blur kernel with various types of neural rendering baseline, such as TensoRF~\cite{chen2022tensorf}, and 3DGS~\cite{kerbl20233d}.
DP-NeRF~\cite{lee2023dp} proposes rigid blur kernel that predict the camera motion during image acquisition process as 3D rigid body motion to preserve the geometric consistency across the scene.
BAD-NeRF~\cite{wang2023bad} and BAD-Gaussians~\cite{zhao2024bad} similarly predict blur kernel as camera motions based on NeRF~\cite{mildenhall2020nerf} and 3DGS~\cite{kerbl20233d}, which assume the simple camera motion and interpolate them between predicted initial and final poses and design simple spline-based methods. 
Deblurring 3DGS~\cite{lee2024deblurring} adjusts Gaussian parameters like rotation and scaling to generate blurry images during training, and BAGS~\cite{peng2024bags} proposes CNN-based multi-scale blur-agonostic degradation kernel with blur masking that indicates the blurred areas.
In this paper, we implement the continuous motion trajectory that forms camera motion blur in 3D using a neural ODE, which fundamentally differentiates our approach from existing methods.
\section{Preliminary}
\label{sec:preliminary}

\subsection{3D Scene Blind Deblurring}
For conventional image blind deblurring~\cite{whyte2012non,chakrabarti2016neural,srinivasan2017light}, the blurring kernel is estimated without any supervision. 
The process to acquire blurry images is achieved by convolving the kernel with sharp images, which takes a fixed grid of size around the pixel location $p$. 
Deblur-NeRF~\cite{ma2022deblurnerf} applies this algorithm to NeRF~\cite{mildenhall2020nerf}, modeling an adaptive sparse kernel for 3D scene representation from blurry images. 
Deblur-NeRF acquires the blurry pixel color $\mathbf{c}_{blur}$ by warping the original input ray into multiple rays that constitute the blur, then determining the pixel color from the colors obtained from these rays:
\begin{equation} \label{eq:1_blur_integraion}
	\mathbf{c}_{blur} = \sum_{i=0}^{N} w_{p}^{i}\mathbf{c}_{p}^{i},~w.r.t.~\sum_{i=0}^{N}w_{p}^{i}=1,
\end{equation}
where $N$ and $i$ respectively denote the number of warped rays for convolution and the correspoding index; $w_{p}$ is the corresponding weight at each ray's location, and $\mathbf{c}_{p}$ represents the pixel color of sharp image. 
%Deblur-NeRF enhances learning efficiency by setting the number of warped rays smaller than the kernel size of 2D convolution. 
In this paper, instead of warping rays, we propose a method to obtain blurry images by applying \cref{eq:1_blur_integraion} to rendered images on estimated camera trajectory.

\subsection{3D Gaussian Splatting}

Unlike ray tracing-based methods~\cite{mildenhall2020nerf,chen2022tensorf,barron2021mipnerf}, 3DGS~\cite{kerbl20233d} is built on a rasterization-based approach with differentiable 3D Gaussians. 
These 3D Gaussians are initialized from a sparse point cloud obtained via a Structure-from-Motion (SfM)~\cite{schonberger2016pixelwise,shan2008highmotiondeblurring} algorithm and are defined as follows: 
\begin{equation} \label{eq:2_gaussian_represent}
	\mathbf{G}(\mathbf{x}) = e^{-\frac{1}{2}(\mathbf{x} - \mu)^{\top}\mathbf{\Sigma}^{-1}(\mathbf{x} - \mu)},
\end{equation}
where $\mathbf{x}\in\mathbb{R}^{3}$ is a point on the Gaussian $\mathbf{G}$ centered at the mean vector $\mu\in\mathbb{R}^{3}$ with an covariance matrix $\mathbf{\Sigma}\in\mathbb{R}^{3\times 3}$. 
The 3D covariance matrix $\mathbf{\Sigma}$ is derived from a learnable scaling vector $\mathbf{s}\in\mathbb{R}^{3}$ and rotation quaternion $\mathbf{q}\in\mathbb{R}^{4}$, from which the scaling matrix $\mathbf{S}\in\mathbb{R}^{3\times 3}$ and rotation matrix $\mathbf{R}\in\mathbb{R}^{3\times 3}$ are obtained and represented as follows: $\mathbf{\Sigma} = \mathbf{R}\mathbf{S}\mathbf{S}^{\top}\mathbf{R}^{\top}$.

For differentiable splatting~\cite{yifan2019differentiable}, the Gaussians in the 3D world coordinate system are projected into the 2D camera coordinate system. 
This projection uses the viewing transformation $\mathbf{W}\in\mathbb{R}^{3\times 3}$ and the Jacobian $\mathbf{J}\in\mathbb{R}^{2\times 3}$ of the affined approximation of the projective transformation to derive the 2D covariance: $\mathbf{\Sigma}^{\textrm{2D}} = \mathbf{J}\mathbf{W}\mathbf{\Sigma}\mathbf{W}^{\top}\mathbf{J}^{\top}$

Each Gaussian includes a set of spherical harmonics (SH) coefficients and an opacity value $\alpha$ to represent view-dependent color $\mathbf{c}$. 
The pixel color $\mathbf{c}_{p}$ is then obtained by applying alpha blending to $\mathcal{N}$ ordered Gaussians:
\begin{equation} \label{eq:3_alpha_blending}
	\mathbf{c}_{p}=\sum_{i \in \mathcal{N}} \mathbf{c}_i \alpha_i \prod_{j=1}^{i-1}\left(1-\alpha_j\right).
\end{equation}

In this paper, we continuously model camera poses to project Gaussians $\mathbf{G}$ onto the 2D camera coordinate system over the exposure time and obtain the final blurry image through a rasterization process with the obtained poses.

\begin{figure*}[t]
	\centering
	\includegraphics[width=\textwidth]{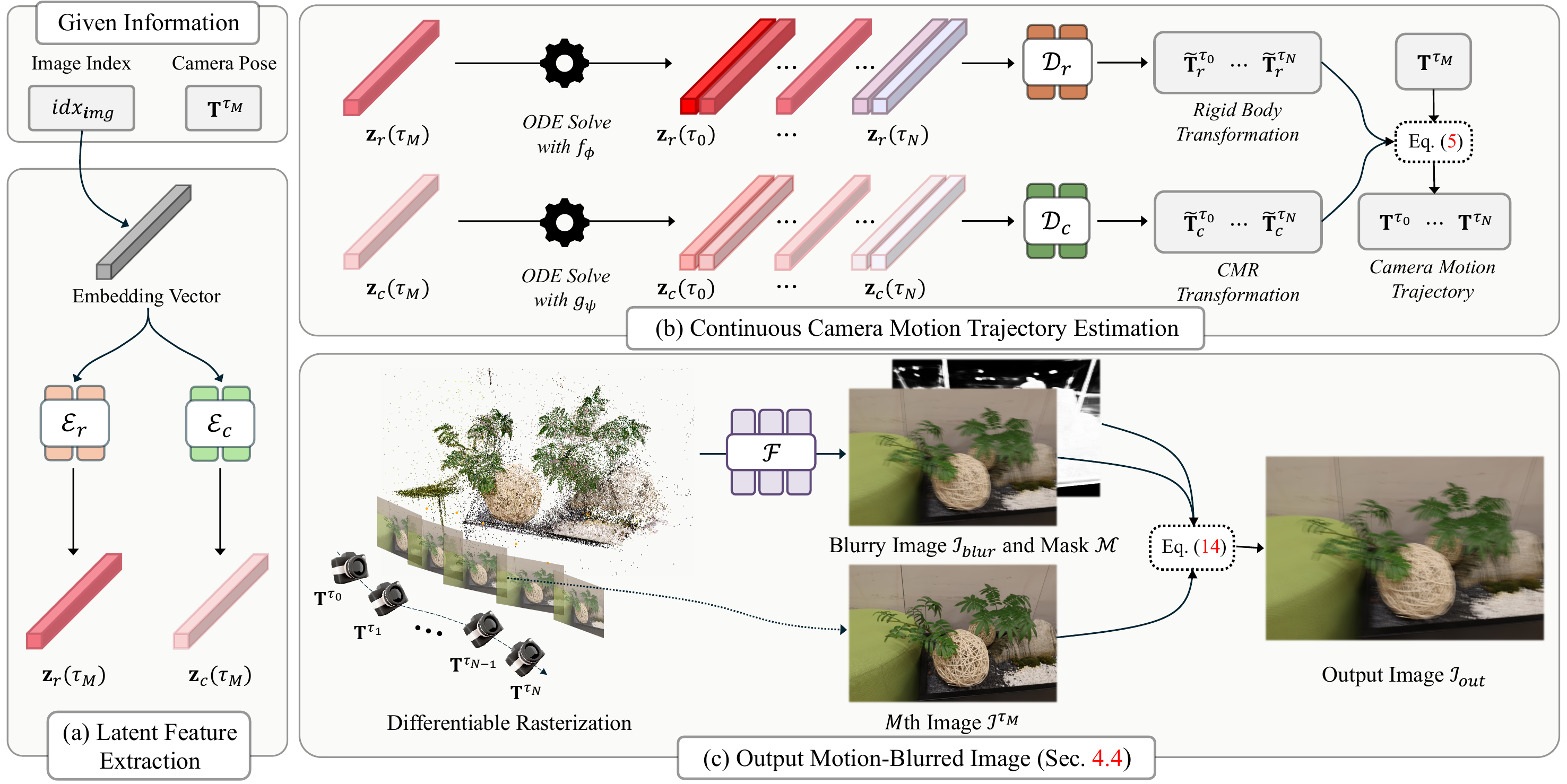}
	\caption{Pipeline of CoMoGaussian. (a) The given image index is embedded and passed through the encoders $\mathcal{E}_{r}$ and $\mathcal{E}_{c}$ to obtain latent features for the rigid body transformation and the CMR transformation. (b) CoMoGaussian solves the ODEs using the latent features and the neural derivatives $f_{\phi}$ and $g_{\psi}$, and obtains $N$ transformed camera poses via \cref{eq:5_trajectory}. (c) Then, $N$ images are rendered through Mip-Splatting~\cite{yu2024mip}, and the output motion-blurred image is obtained through the pixel-wise weighted-sum and scalar pixel mask (\cref{sec:optimize})}
	\label{fig:2_framework}
	\vspace{-3mm}
\end{figure*}

\subsection{Neural Ordinary Differential Equations}

Neural ODEs~\cite{chen2018neural} are first proposed as an approach that interprets neural networks as the derivatives of a ODE systems, where ODEs represent the dynamics inherent to the hidden states. 
Specifically, neural ODEs are utilized to represent parameterized, time-continuous dynamics in the latent space, providing a unique solution given an initial value and numerical differential equation solvers~\cite{lindelof1894application}. 
%Recently, there has been extensive research on neural ODEs. 
%For instance, Latent-ODE~\cite{rubanova2019latent} models the continuous dynamics of irregularly sampled time-series data, while Vid-ODE~\cite{park2021vid} models a smooth latent space by capturing the underlying continuous dynamics of videos.

Neural ODEs model a continuous and differentiable latent state $\mathbf{z}(\tau)$. Within an infinitesimally small step limit $\epsilon$ in the latent space, the local continuous dynamics are modeled as $\mathbf{z}(\tau+\epsilon) = \mathbf{z}(\tau) + \epsilon \cdot \frac{d\mathbf{z}(\tau)}{d\tau}$. The derivative of the latent state, $\frac{d\mathbf{z}(\tau)}{d\tau}$, is represented by a neural network $f(\mathbf{z}(\tau), \tau; \phi)$ parameterized by learnable parameters $\phi$. The latent state at any arbitrary time $\tau_s$ is obtained by solving the ODE from the initial time $\tau_0$:
\begin{equation} \label{eq:4_neural_ode}
	\mathbf{z}\left(\tau_{s}\right)=\mathbf{z}\left(\tau_0\right)+\int_{\tau_0}^{\tau_s} f(\mathbf{z}(\tau), \tau; \phi) d \tau.%=\textrm{ODESolve}\left(\mathbf{z}(t_0), f, t_0, t_s,\phi\right).
\end{equation}
The derivative $f$ modeled by the neural network is expressed as a uniformly Lipschitz continuous non-linear function in $\mathbf{z}$ and $\tau$~\cite{ince1956ordinary}. 
Therefore, the solution obtained through the solver for any given integration interval $(\tau_{i}, \tau_{j})$ is always unique in the integration of the continuous dynamics. 
The simplest method to solve ODE is the Euler method~\cite{euler1845institutionum}, which is a fixed-step-size first-order solver. Additionally, the Runge-Kutta~\cite{kutta1901beitrag} methods are preferred as a higher-order solvers as they offer enhanced stability.

%The derivative $f$ modeled by the neural network is expressed as a uniformly Lipschitz continuous non-linear function in $\mathbf{z}$ and $\tau$~\cite{ince1956ordinary}. 
%Therefore, the solution obtained through the solver for any given integration interval $(\tau_{i}, \tau_{j})$ is always unique in the integration of the continuous dynamics. 
We model the 3D camera trajectory as time-continuous using neural ODEs in the latent space for a continuous representation. 
In our experiments, we adopt the fourth-order Runge-Kutta solver for all experiments, following \cite{chen2018neural}.
\section{Method}
\label{sec:method}
\subsection{CoMoGaussian Framework}

Our goal is to reconstruct a sharp 3D scene using only camera motion-blurred images as input, thereby obtaining deblurred novel view images. 
Inspired by image blind deblurring methods, we follow the approach of Deblur-NeRF~\cite{ma2022deblurnerf}, learning a kernel that intentionally blurs images and excluding this kernel during rendering to produce sharp images. 
Since sharp image supervision is not available during training, we adhere to the fundamental principle of accurately simulating motion-blurred images. 
In this framework, learning to reconstruct blurry images inherently drives the model to capture the underlying sharp image representation.
As illustrated in \cref{fig:1_teaser}, our blurring process consists of camera poses along the camera motion trajectory, generated continuously in time order through neural ODEs~\cite{chen2018neural}. 
Each pose is composed of a rigid body transformation $\tilde{\mathbf{T}}_{r}=[\tilde{\mathbf{R}}_{r}|\tilde{\mathbf{t}}_{r}]$ (\cref{sec:rigid}), which maintains the shape and size of the subject, and a continuous motion refinement (CMR) transformation $\tilde{\mathbf{T}}_{c}=[\tilde{\mathbf{R}}_{c}|\tilde{\mathbf{t}}_{c}]$ (\cref{sec:refine}), which compensates for the limitations of approximating continuous motion using a discrete sum of rigid transformations by introducing learnable adjustments.
Note that $\mathbf{R} \in \mathbb{R}^{3 \times 3}$ and $\mathbf{t} \in \mathbb{R}^{3}$ represent rotation matrix and translation vector, and $\tilde{\mathbf{R}} \in \mathbb{R}^{3 \times 3}$ and $\tilde{\mathbf{t}} \in \mathbb{R}^{3}$ stand for multiplicative offsets of change of the rotation matrix and translation vector, respectively. 
Given the two transformation matrices $\tilde{\mathbf{T}}_{r}=[\tilde{\mathbf{R}}_{r}|\tilde{\mathbf{t}}_{r}]$ and $\tilde{\mathbf{T}}_{c}=[\tilde{\mathbf{R}}_{c}|\tilde{\mathbf{t}}_{c}]$, the subsequent pose $\mathbf{T}^{\tau_s} (0\leq s<N)$ in the camera motion is derived as follows:
\begin{equation} \label{eq:5_trajectory}
	\mathbf{T}^{\tau_s} = \mathbf{T}^{\tau_{M}} \tilde{\mathbf{T}}_{r}^{\tau_s} \tilde{\mathbf{T}}_{c}^{\tau_s},~where~\mathbf{T} = [\mathbf{R}|\mathbf{t}] = \left[\begin{array}{cc} \mathbf{R} & \mathbf{t} \\ 0 & 1 \end{array}\right],
\end{equation}
where $\mathbf{T}^{\tau_M} (M=N/2)$ denotes given calibrated camera pose. 
We assume that calibrated camera pose is located at the midpoint of the camera motion trajectory, as it represents an averaged position influenced by the integration of motion blur over the exposure time.
%where $\mathbf{T}^{\tau_0}=[\mathbf{R}^{\tau_0}|\mathbf{t}^{\tau_0}]$ and  $\mathbf{T}^{\tau_s}=[\mathbf{R}^{\tau_s}|\mathbf{t}^{\tau_s}]$ denote the camera poses at the initial state and the time $\tau_s$. 
By rendering $N$ images from the obtained $N$ camera poses and computing their pixel-wise weighted sum (\cref{sec:optimize}), we obtain the final blurred image. Our framework is shown in \cref{fig:2_framework}.

\subsection{Continuous Rigid Body Motion} \label{sec:rigid}

To design the camera motion trajectory within the exposure time, we apply rigid body transformation in that the shape and size of the object remain unchanged. 
Rigid body transformation requires three components for the unit screw axis $\mathcal{S} = (\hat{\omega}, v)$: the unit rotation axis $\hat{\omega}\in\mathbb{R}^3$, the rotation angle $\theta$ about the axis, and the translation component $v\in\mathbb{R}^{3}$ for translation. 
Considering that the direction and extent of blur vary for each image in a single scene, we embed the image index of the scene to obtain different embedded features for each image as shown in~\cref{fig:2_framework}. 
These features are then passed through an encoder $\mathcal{E}_r$ with a single-layer MLP, transforming them into the latent features $\mathbf{z}_{r}(\tau_0)$, which potentially represent for $\hat{\omega}$, $\theta$, and $v$ of $\mathcal{S}$. 
Then, we model the continuous latent space for the screw axis using a neural ODEs~\cite{chen2018neural} to assign the latent continuity to the screw axis in camera motion trajectory. 
The neural derivative $f$ of the latent features for the screw axis is expressed as $\frac{d\mathbf{z}_{r}(\tau)}{d\tau} = f(\mathbf{z}_{r}(\tau), \tau; \phi)$, and the latent features at an arbitrary time $\tau_s$ can be obtained by ODE solver, numerically integrating $f$ from $\tau_M$ to $\tau_s$:
\begin{equation} \label{eq:6_ode_rigid}
	\mathbf{z}_{r}\left(\tau_{s}\right)=\mathbf{z}_{r}\left(\tau_M\right)+\int_{\tau_M}^{\tau_s} f(\mathbf{z}_{r}(\tau), \tau; \phi) d \tau.
\end{equation}
To obtain $N$ poses along the camera trajectory, we uniformly sample $N$ time points within the fixed exposure time and get $N$ latent features by applying the~\cref{eq:6_ode_rigid} to them. 
We then transform the latent features obtained by the ODE solver into the unit screw axis $\mathcal{S}$ using a single-layer MLP decoder $\mathcal{D}_{r}$. 
In Nerfies~\cite{park2021nerfies} and DP-NeRF~\cite{lee2023dp}, the angular velocity $\omega = \hat{\omega} \theta$ is modeled and then decomposed into the unit rotation axis $\hat{\omega}$ and the rotation angle $\theta$, making the two elements dependent on each other. 
However, since $\theta$ represents the rotation amount about the axis, these two elements should be independent. 
Therefore, we model $\hat{\omega}$ with normalization and $\theta$ independently through the decoder $\mathcal{D}_{r}$:
\begin{equation} \label{eq:7_decode_rigid}
	\mathcal{D}_{r}\left(\mathbf{z}_{r}(\tau)\right) = (\hat{\omega}^{\tau}, \theta^{\tau}, v^{\tau}), \ where\ ||\hat{\omega}^{\tau}|| = 1.
\end{equation}
According to~\cite{lynch2017modernrobotics}, the screw axis $\mathcal{S}^{\tau}=(\hat{\omega}^{\tau}, v^{\tau})$ is a normalized twist, so the infinitesimal transformation matrix $[\mathcal{S}^{\tau}] \in \mathbb{R}^{4 \times 4}$ is represented as follows:
\begin{equation} \label{eq:8_se3}
	[\mathcal{S}^{\tau}] = \left[\begin{array}{cc}
		{[\hat{\omega}^{\tau}]} & v^{\tau} \\
		0 & 0
	\end{array}\right] \in \mathfrak{se}(3),
\end{equation}
where $[\hat{\omega}^{\tau}]\in\mathfrak{so}(3)$ is a $3\times 3$ skew-symmetric matrix of vector $\hat{\omega}^{\tau}$. 
To derive the infinitesimal transformation matrix $[\mathcal{S}^{\tau}] \theta^{\tau} \in \mathfrak{se}(3)$ on the Lie Algebra to the transformation matrix $\mathbf{T}_{r}^{\tau} \in \textit{SE(3)}$ in the Lie Group, we use the matrix exponential $e^{[\mathcal{S}^{\tau}] \theta^{\tau}}$, whose rotation matrix and translation vector are $\mathbf{R}^{\tau}_{r}=e^{[\hat{\omega}^{\tau}] \theta^{\tau}}$ and $\mathbf{t}^{\tau}_{r}=G(\theta^{\tau}) v^{\tau}$, respectively. 
These matrices are expressed as follows using Rodrigues' formula~\cite{rodrigues1816attraction} through Taylor expansion:
\begin{equation}\label{eq:9_SO3}
	e^{[\hat{\omega}^{\tau}] \theta^{\tau}} = I + \sin \theta^{\tau} [\hat{\omega}^{\tau}] + (1 - \cos \theta^{\tau}) [\hat{\omega}^{\tau}]^2 \in \textit{SO(3)},
\end{equation} \vspace{-3mm}
\begin{equation}\label{eq:10_G_theta}
	G(\theta^{\tau}) = I \theta^{\tau} + (1 - \cos \theta^{\tau}) [\hat{\omega}^{\tau}] + (\theta^{\tau} - \sin \theta^{\tau}) [\hat{\omega}^{\tau}]^2.
\end{equation}
Through the whole process, we obtain $N$ continuous rigid body transformation matrices $\mathbf{T}_{r}=e^{[\mathcal{S}] \theta}$, and by multiplying these with the input pose as in \cref{eq:5_trajectory}, we obtain the transformed poses. 
We show whole derivation process of \cref{eq:9_SO3,eq:10_G_theta} in the \textbf{appendix}, referring \cite{lynch2017modernrobotics}.

\subsection{Continuous Motion Refinement} \label{sec:refine}

We predict the continuous camera trajectory using only rigid body transformations; however, according to \cref{eq:1_blur_integraion}, the computation of blurry pixels through numerical integration is inherently discrete. 
In other words, while a blurry pixel should ideally be formed by continuous integration over time, we approximate it using only $N$ discrete samples. 
Consequently, rigid body transformations alone are insufficient to accurately represent the continuous motion of the camera under discretized numerical integration. 
To address this limitation, we introduce the CMR Transformation.
This transformation has higher degrees of freedom from a learning perspective and is simple to implement. 
The CMR transformation optimizes the transformation matrix close to \textit{SE(3)} without explicitly enforcing its constraints. 
Unlike traditional rigid body transformations that strictly adhere to \textit{SE(3)} by parameterizing motion through a screw axis representation $(\hat{\omega}, v)$, the CMR transformation learns a transformation matrix $\tilde{\mathbf{R}}_{c}$ and a translation vector $\tilde{\mathbf{t}}_{c}$ in an unconstrained space. 
%The CMR transformation optimizes the transformation matrix close to \textit{SE(3)} directly without any screw axis components as described in \cref{eq7}. 
%Thus, instead of modeling $\hat{\omega}$ and $v$, this transformation designs a matrix $\tilde{\mathbf{R}}_{c}$ which is regularized close to rotation matrix, and the translation vector $\tilde{\mathbf{t}}_{c}$. 
This process begins by encoding the image index into the latent state $\mathbf{z}_{c}(\tau_M)$ of $\tilde{\mathbf{R}}_{c}$ and $\tilde{\mathbf{t}}_{c}$ using the encoder $\mathcal{E}_{c}$ with a single-layer MLP, similar to \cref{sec:rigid}. 
The latent state $\mathbf{z}_{c}(\tau_s)$ at any arbitrary time $\tau_s$ is obtained using the neural derivative $g$ parameterized by $\psi$ and a solver:
\begin{equation} \label{eq:11_ode_cmr}
	\mathbf{z}_{c}(\tau_s) = \mathbf{z}_{c}(\tau_M) + \int_{\tau_M}^{\tau_s} g(\mathbf{z}_{c}(\tau), \tau; \psi) \, d\tau.
\end{equation}
The latent features are then decoded into rotation matrix component $\mathbf{A}^{\tau}_{c}$ and translation vector $\tilde{\mathbf{t}}^{\tau}_{c}$ through a single-layer MLP decoder $\mathcal{D}_{c}$:
\begin{equation} \label{eq:12_decode_cmr}
	\mathcal{D}_{c}(\mathbf{z}_{c}(\tau)) = (\mathbf{A}_{c}^{\tau}, \tilde{\mathbf{t}}_{c}^{\tau}) \rightarrow \tilde{\mathbf{T}}_{c}^{\tau} = \left[\begin{array}{cc}
		\tilde{\mathbf{R}}_{c}^{\tau} & \tilde{\mathbf{t}}_{c}^{\tau} \\
		0 & 1
	\end{array}\right],
\end{equation}
where $\tilde{\mathbf{R}}^{\tau}_{c} = \mathbf{A}^{\tau}_{c} + \mathbf{I}$ and $\mathbf{I}$ is the identity matrix to initialize $\tilde{\mathbf{R}}^{\tau}_{c}$ to identity transformation. 
Since the CMR transformation is intended to refine the rigid body transformation, it should not significantly affect the rigid body transformation during the initial stages of training. 
Therefore, we initialize the weights of the decoder to approximate $\mathbf{A}_{c}^{\tau}$ to zero matrix using a uniform distribution $\mathcal{U}(-10^{-5}, 10^{-5})$, which means that $\tilde{\mathbf{R}}_{c}^{\tau}$ is initialized close to identity matrix. 
To ensure that $\tilde{\mathbf{R}}_{c}$ remains close to a vaild rotation matrix, we apply explicit regularization, allowing for greater flexibility while maintaining consistency with rigid motion. 
The proposed loss is $\mathcal{L}_{o}$, which ensures the orthogonality condition of rotation matrix: $\mathcal{L}_{o} = \|\tilde{\mathbf{R}}_{c}^{\top} \tilde{\mathbf{R}}_{c} - \mathbf{I}\|_2$. 
Finally, we apply \cref{eq:1_blur_integraion} using $\tilde{\mathbf{T}}_{c}^{\tau_s}$ and $\tilde{\mathbf{T}}_{r}^{\tau_s}$ obtained from \cref{sec:rigid} to get the refined transformed camera pose $\mathbf{T}^{\tau_s}$. 

\subsection{Optimization} \label{sec:optimize}
\paragraph{Pixel-wise Weight and Mask.}
Once the $N$ images along the camera motion trajectory are rendered from the $N$ camera poses, we apply a pixel-wise weighted sum to create the blurry image $\mathcal{I}_{blur}$, following previous research~\cite{ma2022deblurnerf,lee2023dp,lee2024smurf,peng2023pdrf}. 
To satisfy \cref{eq:1_blur_integraion}, we use a shallow CNN $\mathcal{F}$ and a softmax function to compute the pixel-wise weights $\mathcal{P} \in \mathbb{R}^{N \times H \times W \times 3}$ for the resulting images, as follows:
\begin{equation} \label{eq:13_pixel_sum}
	\mathcal{I}_{blur} = \sum_{i=1}^{N}{\mathcal{I}^{\tau_i} \cdot \mathcal{P}^{\tau_i}}, ~ where~\mathcal{P} = \textrm{softmax}(\mathcal{F}(\mathcal{I})),
\end{equation}
where $\mathcal{I}^{\tau_i}$ is the $i$-th image along the camera motion, and $\mathcal{P}^{\tau_i}$ is the pixel-wise weight of $i$-th image. 
Additionally, we adopt per-pixel scalar mask~\cite{peng2024bags} $\mathcal{M}\in\mathbb{R}^{H\times W\times 3}$ to generate final output blurry image $\mathcal{I}_{out}$ by blending the sharp image $\mathcal{I}^{\tau_M}$ at camera pose $\mathbf{T}^{\tau_M}$ and the blurry image $\mathcal{I}_{blur}$ acquired by pixel-wise weight: 
\begin{equation} \label{eq:14_output}
	\mathcal{I}_{out}=(\mathbf{1} - \mathcal{M})\cdot\mathcal{I}^{\tau_M} + \mathcal{M}\cdot\mathcal{I}_{blur}.
\end{equation}
The scalar mask aims to decide whether each pixel is blurry or not, and a small sparsity constraint is applied to the mask to assign a larger weight on $\mathcal{I}^{\tau_M}$, namely mask sparsity loss $\mathcal{L}_{\mathcal{M}}$, which is a mean value of the scalar mask. 
This method allows us to obtain the final precise blurry image, which is then optimized against the ground truth blurry image.

\paragraph{Objective.} We optimize the learning process using the $\mathcal{L}_{1}$ loss and D-SSIM between the generated output blurry image $\mathcal{I}_{out}$ and the ground truth blurry image, similar to 3D-GS~\cite{kerbl20233d}. 
The $\mathcal{L}_{1}$ loss ensures pixel-wise accuracy, while D-SSIM captures perceptual differences. 
Additionally, we apply the regularization loss $\mathcal{L}_{o}$ to regularize the CMR transformation and mask sparsity loss $\mathcal{L}_\mathcal{M}$. The final objective $\mathcal{L}$ is defined as follows:
\begin{equation}
	\mathcal{L} = (1 - \lambda_{c})\mathcal{L}_{1} + \lambda_{c}\mathcal{L}_{\rm{D-SSIM}} + \lambda_{o}\mathcal{L}_{o} + \lambda_{\mathcal{M}}\mathcal{L}_{\mathcal{M}},
\end{equation}
where $\lambda_{c}$ is a factor for balancing $\mathcal{L}_{1}$ and $\mathcal{L}_{\rm{D-SSIM}}$, and $\lambda_{o}$ and $\lambda_{\mathcal{M}}$ are factors for $\mathcal{L}_{o}$ and $\mathcal{L}_{\mathcal{M}}$, respectively.
\begin{figure*}[t]
	\centering
	\includegraphics[width=\linewidth]{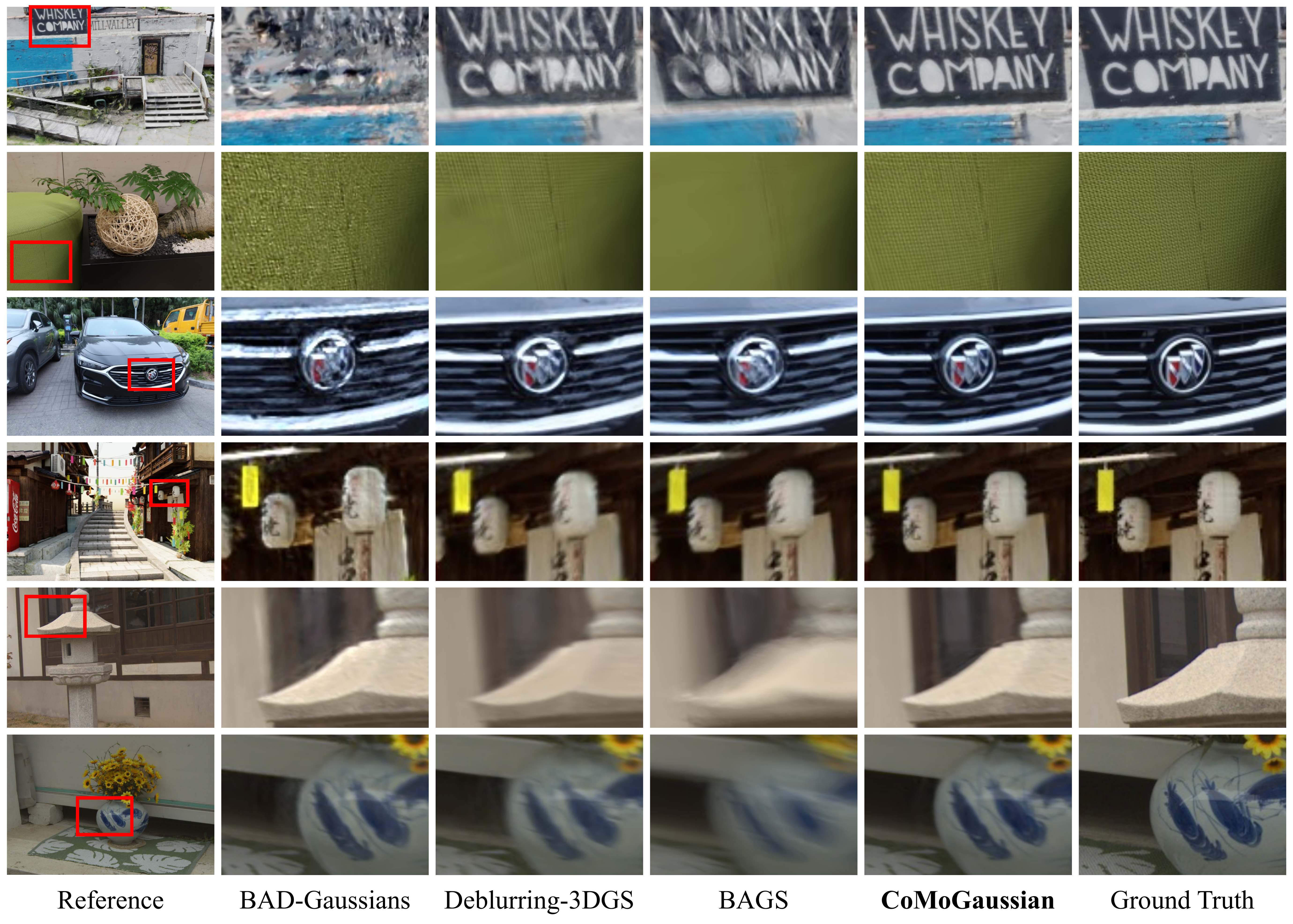}
	\vspace{-7mm}
	\caption{Qualitative comparison on the Deblur-NeRF~\cite{ma2022deblurnerf} synthetic and real-world scenes, and ExbluRF~\cite{lee2023exblurf} real-world scenes.}
	\label{fig:3_comparison}
	\vspace{-4mm}
\end{figure*}

\section{Experiments}
\label{sec:experiments}

\paragraph{Datasets.} 
CoMoGaussian is evaluated using three datasets: the Deblur-NeRF~\cite{ma2022deblurnerf} synthetic dataset, the Deblur-NeRF real-world dataset, and the ExbluRF~\cite{lee2023exblurf} real-world dataset. 
The Deblur-NeRF synthetic dataset consists of 5 scenes generated using Blender~\cite{blender}, where the images are combined in linear RGB space to create the final blurred images.
The Deblur-NeRF real-world dataset includes 10 scenes captured with a CANON EOS RP camera, where the exposure time is manually set to produce blurry images. 
The ExbluRF real-world dataset consists of 8 scenes with challenging camera motion. 
We obtain the camera poses for each image and the initial point clouds by applying COLMAP~\cite{schonberger2016pixelwise,shan2008highmotiondeblurring}. 

\begin{table}[!t] 
	\begin{center}
		\caption{Comparisons on Deblur-NeRF synthetic and real-world scene dataset. ``*'' denotes the results obtained by reproducing the released code. The \colorbox{best!25}{orange} and \colorbox{second!35}{yellow} cells respectively indicate the highest and second-highest value.}
		\vspace{-2mm}
		\resizebox{\columnwidth}{!}{
			\centering
			\setlength{\tabcolsep}{1pt}
			\begin{tabular}{l||c|c|c|c|c|c}
				\toprule 
				
				\multirow{2}{*}{Methods} 			   	& \multicolumn{3}{c|}{~Synthetic Scene~\cite{ma2022deblurnerf}~ }  	   & \multicolumn{3}{c}{~Real-World Scene~\cite{ma2022deblurnerf}~}  	\\ \cmidrule{2-7}
				&~PSNR$\uparrow$~    &~SSIM$\uparrow$~    &~LPIPS$\downarrow$~ &~PSNR$\uparrow$~    &~SSIM$\uparrow$~    &~LPIPS$\downarrow$~     	\\ \midrule \midrule
				Naive NeRF~\cite{mildenhall2020nerf}                            				 & 23.78    & 0.6807   & 0.3362 		& 22.69       & 0.6347      & 0.3687 	 	\\
				Mip-Splatting~\cite{yu2024mip}																			&	23.28			&	0.6765				& 0.2822					& 21.87			& 0.6270	& 0.3066 \\ \midrule 
				%				NeRF+MPR~\cite{zamir2021multi}                           					& 25.11    & 0.7476   & 0.2148 			& 23.38       & 0.6655      & 0.3140      	\\ \midrule
				Deblur-NeRF~\cite{ma2022deblurnerf}                          			& 28.77    & 0.8593   & 0.1400 		 & 25.63       & 0.7645      & 0.1820 	\\
				%				PDRF-10*~\cite{peng2023pdrf}                         					   & 28.86   & \cellcolor{second!35}0.8795   & 0.1139 		 		  & 25.90       & 0.7734      & 0.1825        				\\
				%				BAD-NeRF*~\cite{wang2023bad}										   & 27.32   & 0.8178   & 0.1127 		 		  & 22.82       & 0.6315      & 0.2887        			\\
				DP-NeRF~\cite{lee2023dp}                        						  & \cellcolor{second!35}29.23    & 0.8674   & 0.1184 		& 25.91   	 & 0.7751      & 0.1602   \\ \midrule
				%				DeblurGS*~\cite{oh2024deblurgs}									& 20.22		& 0.5454	& 0.1042			& 20.48		& 0.5813		& 0.1186 \\
				BAD-Gaussians*~\cite{zhao2024bad}                          			& 22.01   & 0.6377   & \cellcolor{second!35}0.1001 		 & 21.69       & 0.6471      & 0.1262 	\\
				Deblurring 3DGS~\cite{lee2024deblurring}~                        				& 28.24    & 0.8580   & 0.1051 		 & 26.61       & 0.8224      & 0.1096 	\\ 
				BAGS~\cite{peng2024bags}                          				& 27.34    & 0.8353   & 0.1116 		 & \cellcolor{second!35}26.70       & \cellcolor{second!35}0.8237      & \cellcolor{second!35}0.0956 	\\ \midrule \midrule
				\textbf{CoMoGaussian}											                        & \cellcolor{best!25}31.02    & \cellcolor{best!25}0.9167   & \cellcolor{best!25}0.0492 		 & \cellcolor{best!25}27.85       & \cellcolor{best!25}0.8431      & \cellcolor{best!25}0.0822	\\ \bottomrule
			\end{tabular}
		}
		\label{tab:1_comparison_deblurnerf}
	\end{center}
	\vspace{-5mm}
\end{table}

\begin{table}[!t]
	\centering
	\caption{Comparisons on ExbluRF real-world scene dataset.}
	\vspace{-2mm}
	\resizebox{0.9\columnwidth}{!}{
		\setlength{\tabcolsep}{7pt} 
		\renewcommand{\arraystretch}{1.0}
		\scriptsize 
		\begin{tabular}{l||c|c|c}
			\toprule
			\multirow{2}{*}{Methods} & \multicolumn{3}{c}{~ExbluRF Real-World Scene~\cite{wu2022dof}~} \\ \cmidrule{2-4}
			& PSNR$\uparrow$ & SSIM$\uparrow$ & LPIPS$\downarrow$ \\ \midrule \midrule
			Mip-Splatting~\cite{yu2024mip} 					& 24.52 & 0.5662 & 0.6003 \\ \midrule
			ExbluRF~\cite{lee2023exblurf}~ 								&  23.45 & 0.5495  & 0.3805  \\ \midrule
			BAD-Gaussians~\cite{zhao2024bad}~ 				&  26.83 & 0.6625  & \cellcolor{second!35}0.3221  \\
			Deblurring 3DGS~\cite{lee2024deblurring}~ & \cellcolor{second!35}27.36 & \cellcolor{second!35}0.6795 & 0.3989 \\ 
			BAGS~\cite{peng2024bags} 							& 24.70 & 0.5843 & 0.5278 \\ \midrule \midrule
			\textbf{CoMoGaussian} 														& \cellcolor{best!25}30.15 & \cellcolor{best!25}0.7559 & \cellcolor{best!25}0.3107 \\ \bottomrule
		\end{tabular}
	}
	\label{tab:2_comparison_exblurf}
\end{table}

\subsection{Novel View Synthesis Results}
For quantitative results, we evaluate CoMoGaussian using three metrics: peak signal-to-noise ratio (PSNR), structural similarity index measure (SSIM), and learned perceptual image patch similarity (LPIPS). 
We compare our method with both ray-based~\cite{mildenhall2020nerf,ma2022deblurnerf,lee2023dp} and rasterization methods~\cite{yu2024mip,zhao2024bad,lee2024deblurring,peng2024bags}. 
%Note that the results for BAD-GS~\cite{zhao2024bad} are obtained by re-running the released code ourselves. 
The overall results for all datasets are shown in \cref{tab:1_comparison_deblurnerf,tab:2_comparison_exblurf}, demonstrating that our method achieves superior performance compared to all other methods. 
%Specifically for the LPIPS, CRiM-GS shows an improvement of approximately 52\% for synthetic scenes and 33\% for real-world scenes than the state-of-the-art, and the best LPIPS performance across all scenes not only for synthetic dataset but also real-world dataset. 
Additionally, benefiting from using Mip-Splatting~\cite{yu2024mip} as our backbone, our proposed method ensures fast training and rendering speeds. 
Although BAD-Gaussians~\cite{zhao2024bad} achieves relatively good LPIPS scores on ExbluRF dataset, their PSNR and SSIM scores are lower, indicating that their camera poses are not properly optimized. 
%More details about pose optimization are in the \textbf{appendix}.

%\begin{table}[t] 
%	\begin{center}
%		\caption{Ablation study on our main components.}
%		\resizebox{\columnwidth}{!}{
%			\centering
%			\setlength{\tabcolsep}{1pt}
%			\begin{tabular}{ccc|ccc}
%				\toprule 
%				\multirow{2}{*}{~Rigid Motion~} & \multirow{2}{*}{~CMR~} & \multirow{2}{*}{~Orth. Loss~} & \multicolumn{3}{c}{Real-World Scene~\cite{ma2022deblurnerf}}  	\\ \cmidrule{4-6}
%				&&						 &~PSNR$\uparrow$~    &~SSIM$\uparrow$~    &~LPIPS$\downarrow$~     	\\ \midrule \midrule
%				&&																					&21.87&0.6270&0.3066 \\
%%				\ding{51} &&																	&25.72&0.7986&0.1239 \\
%%				\ding{51}&&														&27.16&0.8204&0.1080 \\
%				\ding{51}&&									&27.19&0.8249&0.1005 \\
%				\ding{51}&\ding{51}&														&25.92&0.8092&0.1117 \\
%%				\ding{51}&\ding{51}&\ding{51}									&26.02&0.8120&0.1061 \\
%%				\ding{51}&\ding{51}&\ding{51}						&27.52&0.8343&0.0965 \\
%				\ding{51}&\ding{51}&\ding{51}		&\textbf{27.85}&\textbf{0.8431}&\textbf{0.0822} \\ \bottomrule
%			\end{tabular}
%			\label{tab:3_ablation}
%		}
%	\end{center}
%	\vspace{-4mm}
%\end{table}

To qualitatively evaluate the results, we visualize rendering results of several scenes and compare to other methods as shown in \cref{fig:3_comparison}. 
Our method shows superior qualitative results compared to the state-of-the-art rasterization-based methods such as BAD-Gaussians~\cite{zhao2024bad}, BAGS~\cite{peng2024bags}, and Deblurring 3DGS~\cite{lee2024deblurring}. 
The bottom two rows of \cref{fig:3_comparison} stand for visualizations of ExbluRF~\cite{lee2023exblurf} \textsc{Stone Lantern} and \textsc{Sunflowers} scenes, where our CoMoGaussian best preserves the edge details of the ground truth, even when given input images with extreme motion blur.

\subsection{Ablation Study}
To thoroughly demonstrate the effectiveness of rigid body and CMR transformations, we conduct ablation experiments as shown in \cref{tab:3_ablation}. 
Additionally, we perform further ablation studies for our continuous modeling, presented in \cref{tab:4_ablation_ode}.
All the experiments for \cref{tab:3_ablation,tab:4_ablation_ode} are conducted on Deblur-NeRF~\cite{ma2022deblurnerf} real-world scenes.
Additional ablation experiments are provided in the \textbf{appendix}.

\begin{table}[t] 
	\begin{center}
		\caption{Ablation study on our main components.}
		\vspace{-2mm}
		\resizebox{0.8\columnwidth}{!}{
			\centering
			\setlength{\tabcolsep}{1pt}
			\begin{tabular}{ccc|ccc}
				\toprule 
				~Rigid Motion~ & ~CMR~ & ~Orth. Loss~ 	   &  ~PSNR$\uparrow$~    &~SSIM$\uparrow$~    &~LPIPS$\downarrow$~ 	\\ \midrule \midrule
				&&																					&21.87&0.6270&0.3066 \\
				\ding{51}&&									&27.19&0.8249&0.1005 \\
				\ding{51}&\ding{51}&														&27.74&0.8402&0.0878 \\
				\ding{51}&\ding{51}&\ding{51}		&\textbf{27.85}&\textbf{0.8431}&\textbf{0.0822} \\ \bottomrule
			\end{tabular}
			\label{tab:3_ablation}
		}
	\end{center}
	\vspace{-6mm}
\end{table}
\vspace{-2mm}
\paragraph{Ablation on Transformations.} 
We analyze the impact of the main components of CoMoGaussian: the rigid body transformation and the CMR transformation. 
The baseline in \cref{tab:3_ablation} is Mip-Splatting~\cite{yu2024mip}, and when pixel-wise weights are not applied, the $N$ rendered images are averaged to produce a blurry image. 
We observe that incorporating the rigid body transformation alone into the baseline results in significantly improved performance across all metrics compared to the baseline. 
Furthermore, adding the CMR transformation to the rigid body motion, which compensates for the discretized integration, consistently enhances performance across all metrics.
Finally, applying the orthogonality loss $\mathcal{L}_{o}$, a component of the CMR transformation, leads to further improvements in LPIPS.

%\paragraph{Ablation on Pixel-wise Weight and Mask. }
%The pixel-wise weight, inspired by traditional ray-based 3D scene deblurring methods~\cite{lee2023dp,ma2022deblurnerf,peng2024bags} and blind image deblurring techniques~\cite{son2021pvd,zamir2021multi}, differs from the conventional approach of simply averaging multiple images to generate a blurry image. 
%As shown in \cref{tab:3_ablation}, incorporating this method consistently enhances performance compared to not using it.
%This improvement suggests that assigning appropriate weights to each pixel location allows for a more accurate reconstruction of motion-blurred image, ultimately surpassing existing state-of-the-art methods.
%Additionally, applying a per-pixel scalar mask enables the model to learn which pixels are blurry and which are not, leading to consistently improved performance, particularly in LPIPS.
%\vspace{-4mm}
\paragraph{Ablation on Neural ODE.}
To validate the effectiveness of implementing rigid body and CMR transformations continuously in latent space, we conduct ablation studies using various approaches, as shown in \cref{tab:4_ablation_ode}.
For linear interpolation and cubic B-spline, we adopt the implementations from BAD-Gaussians~\cite{zhao2024bad}. 
Since these methods operate in 3D physical space rather than latent space, they exhibit lower performance compared to other latent space-based methods.

For latent space-based methods, we test MLP and GRU~\cite{cho2014learning} as alternatives to neural ODEs.
MLP-based method replaces the neural ODE by extracting $N$ latent features from the latent feature of $\tau_{M}$ using a two-layer MLP.
GRU-based method replaces the neural ODE with GRU cells to generate latent features.
Our neural ODE-based approach consistently outperforms both methods.
Notably, in our neural ODE framework, we extract features for all time points by propagating forward ($\tau_{M}\rightarrow\tau_{N}$) and backward ($\tau_{M}\rightarrow\tau_{0}$) from the midpoint $\tau_{M}$ of the exposure time. 
Since all these processes share the same neural derivative, they remain within the same function space.
However, in the GRU-based approach, forward and backward time features require separate GRU cells, making it theoretically less suitable than our neural ODE-based method.

\begin{table}[t] 
	\begin{center}
		\caption{Ablation study on the type of camera motion estimator.}
		\vspace{-2mm}
		\resizebox{\columnwidth}{!}{
			\centering
			\setlength{\tabcolsep}{1pt}
			\begin{tabular}{l|cc|ccc}
				\toprule 
				Methods & ~Physical Space~ & ~Latent Space~     & ~PSNR$\uparrow$~    &~SSIM$\uparrow$~    &~LPIPS$\downarrow$~    \\ \midrule \midrule
				Linear Interpolation~\cite{zhao2024bad}~ & \ding{51} & & 20.97 & 0.6085 & 0.1461 \\
				Cubic B-Spline~\cite{zhao2024bad}~ & \ding{51} & & 21.69 & 0.6471 & 0.1262 \\ \midrule
				MLP & &\ding{51} & 27.43 & 0.8295 & 0.1010 \\
				GRU & &\ding{51} & 27.40 & 0.8302 & 0.0988 \\
				Neural ODE & &\ding{51} & \textbf{27.85} & \textbf{0.8435} & \textbf{0.0822} \\ \bottomrule
			\end{tabular}
			\label{tab:4_ablation_ode}
		}
	\end{center}
	\vspace{-6mm}
\end{table}

\begin{table}[h] 
	\begin{center}
		\caption{Experimental results on NeRF-LLFF~\cite{mildenhall2019local,mildenhall2020nerf} dataset.}
		\vspace{-2mm}
		\resizebox{0.8\columnwidth}{!}{
			\centering
			\setlength{\tabcolsep}{7pt}
			\begin{tabular}{l|ccc}
				\toprule 
				Methods					 &~PSNR$\uparrow$~    &~SSIM$\uparrow$~    &~LPIPS$\downarrow$~     	\\ \midrule \midrule
				Mip-Splatting~\cite{yu2024mip}									& \cellcolor{best!25}27.71 & \cellcolor{best!25}0.8685  & \cellcolor{second!35}0.0494 \\\midrule
				Deblurring 3DGS~\cite{lee2024deblurring}														& 17.42 &0.5075&0.3187 \\
				BAGS~\cite{peng2024bags}														& 26.35 & 0.8490 & 0.0528 \\
				\textbf{CoMoGaussian}		&\cellcolor{second!35}27.56 &\cellcolor{second!35}0.8669&\cellcolor{best!25}0.0457 \\ \bottomrule
			\end{tabular}
			\label{tab:5_sharp}
		}
	\end{center}
	\vspace{-8mm}
\end{table}

\paragraph{Experiments on Sharp Images.}
To evaluate the generalization capability of CoMoGaussian, we conduct experiments on the NeRF-LLFF~\cite{mildenhall2019local,mildenhall2020nerf} dataset, a forward-facing scenes consisting of sharp images, with the results presented in \cref{tab:5_sharp}. CoMoGaussian performs comparably to Mip-Splatting~\cite{yu2024mip}, demonstrating that our method is not limited to motion-blurred images. Additionally, we evaluate the performance of other methods, Deblurring 3DGS~\cite{lee2024deblurring} and BAGS~\cite{peng2024bags}. They struggle to optimize certain scenes and fall short of CoMoGaussian in quantitative performance. This highlights the superior generalization ability of our approach compared to existing methods.

%\begin{table}[t] 
%	\begin{center}
%		\caption{Comparisons on synthetic and real-world dataset. We evaluate the performance on three metrics (PSNR, SSIM, LPIPS).}
%		\resizebox{\columnwidth}{!}{
%			\centering
%			\setlength{\tabcolsep}{1pt}
%			\begin{tabular}{c|cc|ccc}
%				\toprule 
%				\multirow{2}{*}{Method} & Physical & Latent     & \multicolumn{3}{c}{Real-World Scene}  	\\ \cmidrule{4-6}
%				&Space&Space&~PSNR$\uparrow$~    &~SSIM$\uparrow$~    &~LPIPS$\downarrow$~     	\\ \midrule \midrule
%				Linear Spline & & & & & \\
%				Cubic Spline & & & & & \\
%				MLP & & & & & \\
%				GRU & & & & & \\
%				Neural ODE & & & & & \\
%			\end{tabular}
%		}
%	\end{center}
%	\vspace{-4mm}
%	\label{tab:ablation}
%\end{table}

\subsection{Camera Trajectory Visualization} 
\begin{figure}[t]
	\centering
	\includegraphics[width=\linewidth]{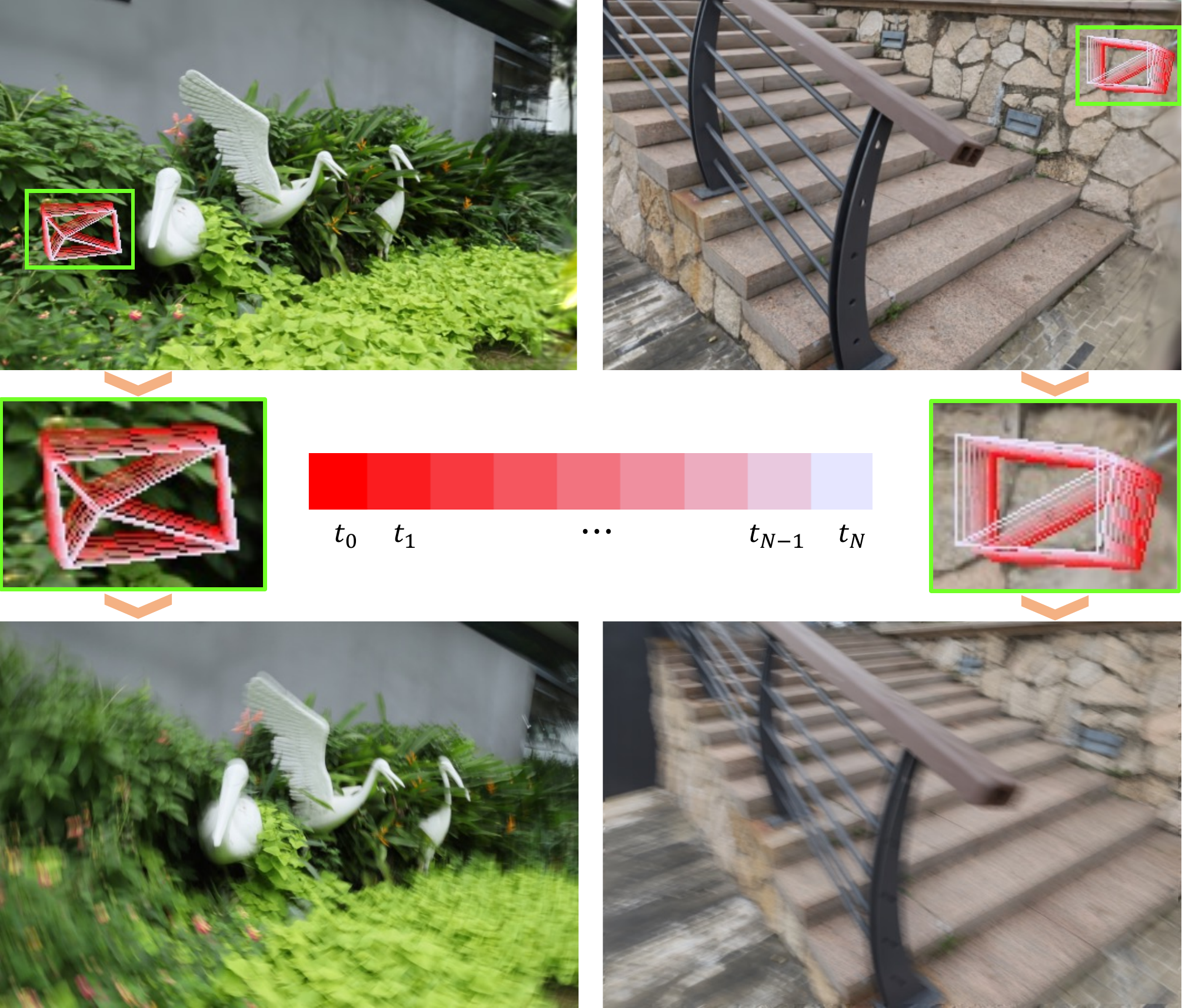}
	\caption{Camera Trajectory Visualization. Red cones stand for camera poses, and the images below are the output blurry images.}
	\label{fig:4_camera_vis}
	\vspace{-2mm}
\end{figure}

We visualize the continuous camera trajectories for the \textsc{Heron} (left) and \textsc{Stair} (right) scenes from the Deblur-NeRF real-world dataset, as shown in \cref{fig:4_camera_vis}. 
The camera trajectory for a single motion-blurred image is represented by colored cones, with the cone's color gradually transitioning from red to light purple as time progresses from $t_{0}$ to $t_{N}$. 
The visualized trajectories confirm that the camera paths generated by CoMoGaussian are smoothly continuous, validating that the continuous transformations described in \cref{sec:rigid,sec:refine} function as intended. 
Additionally, the output blurry images in \cref{fig:4_camera_vis} align with the visualized camera poses, accurately reflecting both the positions and directions of the camera. 
Specifically, even though the camera path in the \textsc{Heron} scene on the left is nonlinear, CoMoGaussian accurately predicts this path and generates a precise blurry image.
Visualizations of camera motion for various types of motion blur, including extreme and moderate motion blur, are provided in the \textbf{appendix}.

%\subsection{Rolling Shutter Effect Compensation}\label{subsec:rolling_shutter}
%To demonstrate the effectiveness of the adaptive distortion-aware transformation in correcting nonlinear distortions, we conducted experiments on a dataset with rolling shutter effects~\cite{seiskari2024gaussian}. As shown in~\cref{fig:rolling_shutter}, while 3DGS~\cite{kerbl20233d} suffers from significant geometric distortions, our method produces results closely matching the ground truth. This capability stems from the inherent flexibility of the adaptive distortion-aware transformation, which can accommodate both linear transformations such as affine transformations and nonlinear ones such as deformable transformations. Thus, as shown in~\cref{tab:ablation}, combining our proposed transformation with rigid body transformations improves performance by addressing the subtle distortions present in motion-blurred images, effectively enhancing the accuracy of the results.
%
%
%
%\begin{figure}[t]
%	\centering
%	\includegraphics[width=\linewidth]{figure/rolling_shutter.pdf}
%	\caption{Rendering Results on Rolling Shutter Effect Dataset~\cite{seiskari2024gaussian}}
%	\label{fig:rolling_shutter}
%\end{figure}
\section{Limitation and Future Work}
\label{sec:limitation}
While our CoMoGaussian achieves both high speed and high-quality results, it lacks the ability to distinguish between moderate and severe blur. 
In other words, moderate blur can be effectively modeled with fewer warped camera poses, whereas severe blur benefits from a greater number of warped poses for more accurate representation. 
Therefore, adaptively determining the number of warped poses based on the degree of blur could improve training efficiency and mitigate potential overfitting issues.
\section{Conclusion}
\label{sec:conclusion}

We propose CoMoGaussian, a novel method for reconstructing sharp 3D scenes from blurry images caused by camera motion. 
We apply a rigid body transformation and further enhance it with CMR transformation to compensate discretized numerical integration. 
These transformations are continuously modeled using neural ODEs, capturing continuous camera motion trajectories. 
Drawing inspiration from prior 3D scene deblurring research, we incorporate a pixel-wise weighting strategy with a lightweight CNN. 
CoMoGaussian surpasses state-of-the-art methods in 3D scene deblurring, with extensive experiments validating the effectiveness of each component.

\paragraph{Acknowledgements.}
This work was supported by the Yonsei Signature Research Cluster Program of 2024 (2024-22-0161), the National Research Foundation of Korea (NRF) grant funded by the Korean government (MSIT)(No. RS-2024-00340745), and the National Research Foundation of Korea (NRF) grant funded by the Korea government (MSIT)(No. RS-2024-00423362). 
{
	\small
	\bibliographystyle{ieeenat_fullname}
	\bibliography{main}
}
\clearpage
\begin{center}
	\textbf{\Large{Appendix}}
	\vspace{5mm}
\end{center}

\section{Implementation Details}
CoMoGaussian is trained for 40k iterations based on Mip-Splatting~\cite{yu2024mip}. 
We set the number of poses $N$ that constitute the continuous camera trajectory to 9, which indicates that $M$ is 5. 
The embedding function of \cref{sec:rigid} is implemented by $\mathtt{nn.Embedding}$ of PyTorch, and the embedded features have the sizes of hidden state of 64. 
The sizes of hidden state of the single-layer encoders $\mathcal{E}_{r}$, $\mathcal{E}_{c}$, and the single-layer decoders $\mathcal{D}_{r}$, $\mathcal{D}_{c}$ are also 64.
$\mathcal{D}_{r}$ consists of three head MLPs that extract the screw axis parameters $\omega$ and $v$, along with $\theta$, while $\mathcal{D}_{c}$ comprises two head MLPs that extract $\mathbf{A}_{c}$ and $\mathbf{t}_{c}$ as described in \cref{sec:refine}.
The neural derivative function $f$ consists of two parallel single MLP layers, where one is designated for the rotation component and the other for the translation component, where $f$ and $g$ share the learnable parameters. 
To ensure nonlinearity in the camera motion within the latent space, we apply the $\mathtt{ReLU}$ activation function to each layer. 
The CNN $\mathcal{F}$ consists of two convolutional layers with 32 channels with kernel size of 5$\times$5 for the first layer and 3$\times$3 for the rest one. 
The pixel-wise weights are obtained by applying a pointwise convolutional layer to the output of $\mathcal{F}$, and the scalar mask $\mathcal{M}$ is obtained by averaging it to the batch axis and applying another pointwise convolution to the output of $\mathcal{F}$.
For first 1k iterations, Gaussian primitives are roughly trained without rigid body transformation and CMR transformation. 
After 1k iterations, those transformations start to be trained without the pixel-wise weight and the scalar mask to allow the initial camera motion path to be sufficiently optimized. 
After 3k iterations, the pixel-wise weight and the scalar mask start training.
We set \(\lambda_{c}\), \(\lambda_{o}\), and \(\lambda_{\mathcal{M}}\) to 0.3, \(10^{-4}\), and \(10^{-3}\) respectively for the objective function. 
All experiments are conducted on a single NVIDIA RTX 4090 GPU.

\section{Additional Ablation Study}

\begin{figure*}[t]
	\centering
	\includegraphics[width=\linewidth]{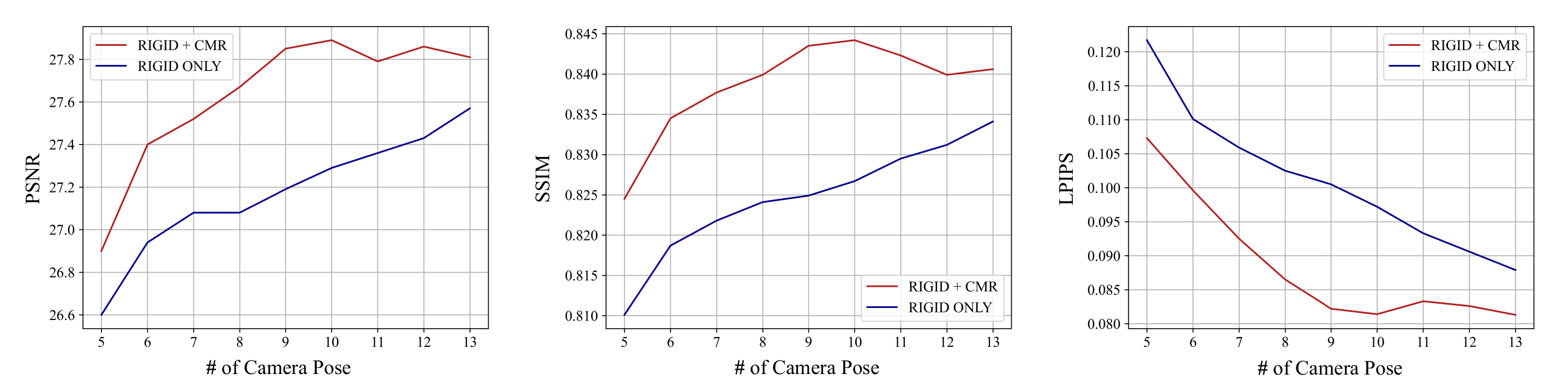}
	\caption{Performance variation based on the number of poses $N$ used to construct the camera motion trajectory.}
	\label{fig:1_num_pose}
\end{figure*}

\paragraph{Number of Poses on Camera Motion.}
We conduct an ablation study on the number of camera poses used to model continuous camera motion, with the results presented in \cref{tab:1_num_pose}. 
To validate the effectiveness of the CMR transformation, we perform two separate experiments: one using only the rigid body transformation described in \cref{sec:rigid}, and the other incorporating both the rigid body transformation and the CMR transformation discussed in \cref{sec:refine}. 
The trends in the three evaluation metrics with respect to different values of $N$ in \cref{tab:1_num_pose} are visualized in \cref{fig:1_num_pose}.

\begin{table}[t]
	\caption{Experimental results based on the number of poses $N$ along the camera motion trajectory.}
	\vspace{-3mm}
	\begin{center}
		\resizebox{\columnwidth}{!}{
			\centering
			\setlength{\tabcolsep}{4pt}
			\begin{tabular}{l||c|c|c|c|c|c}
				\toprule 
				
				\multirow{2}{*}{\# Cam.} 			  	& \multicolumn{3}{c|}{Rigid Only (\cref{sec:rigid})}  	   & \multicolumn{3}{c}{Rigid (\cref{sec:rigid}) + CMR (\cref{sec:refine})}  	\\ \cmidrule{2-7}
				&  ~PSNR$\uparrow$~    &~SSIM$\uparrow$~    &~LPIPS$\downarrow$~ &~PSNR$\uparrow$~    &~SSIM$\uparrow$~    &~LPIPS($\downarrow$)~     	\\ \midrule \midrule
				5	&  26.60 & 0.8101 & 0.1217 & 26.90 & 0.8245 & 0.1073	\\
				6	& 26.94 & 0.8187 & 0.1101 & 27.40 & 0.8345 & 0.0996	\\
				7	& 27.08 & 0.8218 & 0.1059 & 27.52 & 0.8377 & 0.0925	\\
				8	& 27.08 & 0.8241 & 0.1025 & 27.67 & 0.8399 & 0.0865	\\
				9	& 27.19 & 0.8249 & 0.1005 & \cellcolor{second!35}27.85 & \cellcolor{second!35}0.8435 & 0.0822	\\
				10	& 27.29 & 0.8267 & 0.0972 & \cellcolor{best!25}27.89 & \cellcolor{best!25}0.8442 & \cellcolor{second!35}0.0814	\\
				11	& 27.36 & 0.8295 & 0.0933 & 27.79 & 0.8423 & 0.0833	\\
				12	&\cellcolor{second!35} 27.43 & \cellcolor{second!35}0.8312 & \cellcolor{second!35}0.0906 & 27.86 & 0.8399 & 0.0826	\\
				13	& \cellcolor{best!25}27.57 & \cellcolor{best!25}0.8341 & \cellcolor{best!25}0.0879 & 27.81 & 0.8406 & \cellcolor{best!25}0.0813	\\ \bottomrule
			\end{tabular}
		}
	\end{center}
	\label{tab:1_num_pose}
	\vspace{-7mm}
\end{table}

When using only the rigid body transformation, performance improves almost linearly as $N$ increases across all metrics. 
Since a motion-blurred image is fundamentally the result of integrating sharp images over time, increasing $N$ allows for a finer discretization of the continuous camera trajectory, leading to better performance. 
However, numerical integration is inherently discrete, even when the camera motion itself is continuous, which explains why a larger $N$ results in improved accuracy.

Nevertheless, increasing $N$ also leads to a significant rise in computational complexity during training, as each motion-blurred image requires rendering $N$ sharp images. 
This creates a trade-off: smaller values of $N$ suffer from discretization artifacts, while larger values of $N$ lead to increased computational costs due to multiple renderings. 
To address this, we introduce the CMR transformation, which adds a small degree of flexibility to the transformation matrix, effectively compensating for the limitations of discretized integration at smaller $N$.

As shown in \cref{tab:1_num_pose} and \cref{fig:1_num_pose}, when CMR is applied alongside the rigid body transformation, performance saturates at $N=9$ across all metrics, even surpassing the performance of the rigid body transformation alone at $N=13$. 
This demonstrates that the CMR transformation mitigates the limitations of rigid body transformations at lower $N$ and justifies its inclusion in our framework.

\begin{figure}[t]
	\centering
	\includegraphics[width=\linewidth]{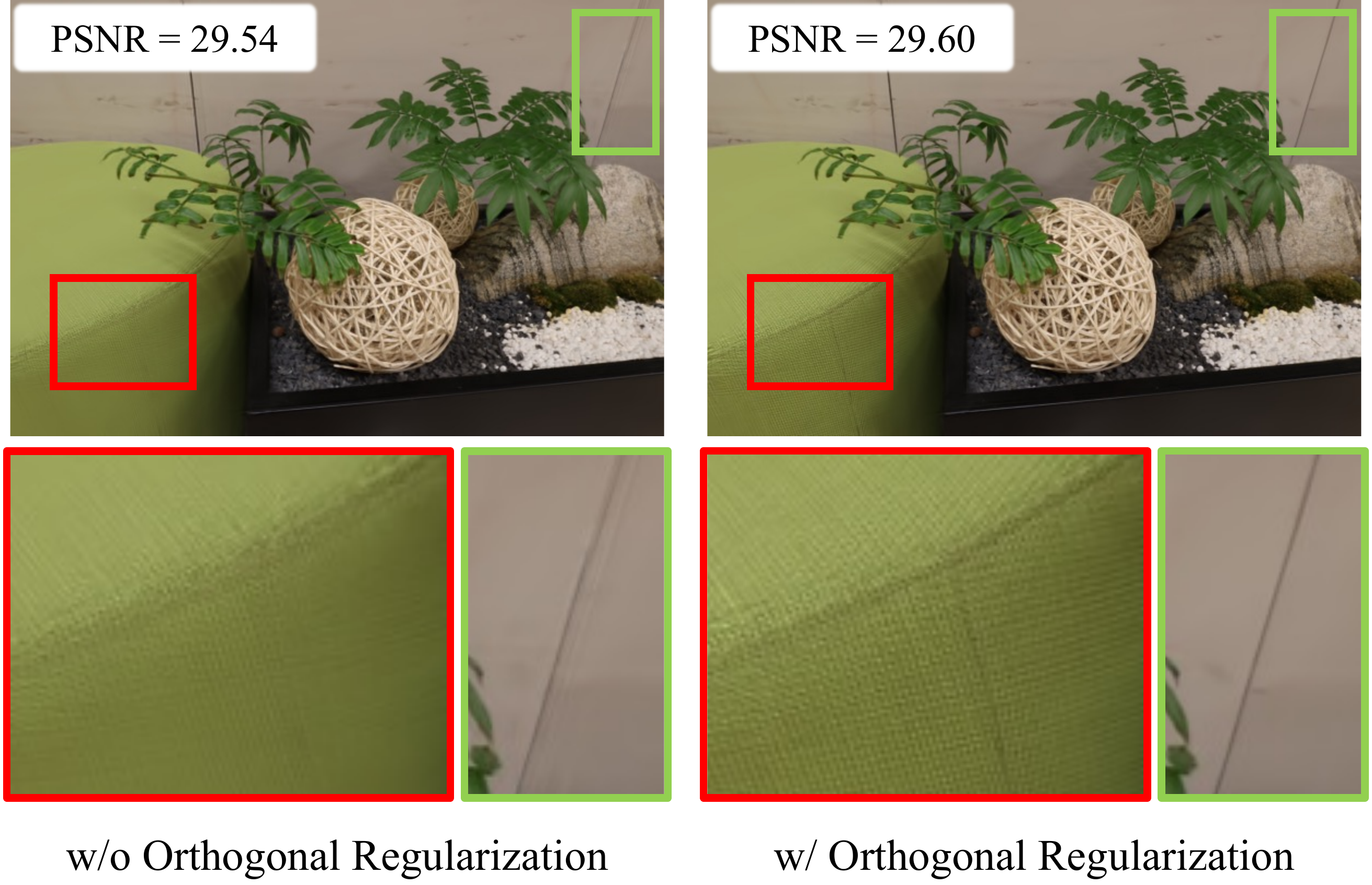}
	\caption{Qualitative comparison of results with and without orthogonal regularization. Despite similar PSNR values, the results with regularization capture more fine details.}
	\label{fig:2_orthogonal}
\end{figure}

\paragraph{Orthogonal Regularization. }
We conduct a qualitative ablation study on the regularization loss for the orthogonality condition introduced in \cref{sec:refine} of the main paper. 
Although the impact of this loss may appear minor in \cref{tab:3_ablation} of the main paper, a qualitative comparison reveals a noticeable difference.

As shown in \cref{fig:2_orthogonal}, including this regularization results in significantly improved visual quality compared to when it is omitted. 
Without regularization, the $3\times 3$ linear transformation matrix is learned without constraints on shearing and scaling, which leads to unintended distortions.
When such unconstrained affine transformations are included, the overall scene structure may still be well captured, but finer details tend to be lost.

By enforcing the orthogonality condition, we restrict the transformation matrix to allow only minimal deviations, preventing excessive distortions. 
As demonstrated in \cref{fig:2_orthogonal}, while the quantitative performance remains similar, the model captures finer details more effectively, highlighting the importance of this regularization.

\section{Difference from SMURF~\cite{lee2024smurf}}
In this section, we compare our approach with SMURF, a methodology for handling the continuous dynamics of camera motion blur. 
SMURF utilizes neural ODEs to warp a given input ray into continuous rays that simulate camera motion. 
However, its continuous dynamics are applied only in the 2D pixel space, lacking the inclusion of higher-dimensional camera motion in 3D space. 
Additionally, as SMURF is implemented on Tensorial Radiance Fields (TensoRF)~\cite{chen2022tensorf}, a ray tracing-based method, it exhibits relatively slower training and rendering speeds.

In contrast, our model uses neural ODE to obtain the 3D camera poses which constitue the camera motion trajectory. 
Our approach incorporates higher-dimensional information compared to SMURF by operating directly in 3D space rather than the 2D pixel space. 
Furthermore, as our method is implemented on Mip-Splatting~\cite{yu2024mip}, a rasterization-based method, it ensures faster training and rendering speeds than SMURF.

\section{Camera Pose Visualization}

\begin{figure}[t]
	\centering
	\includegraphics[width=\linewidth]{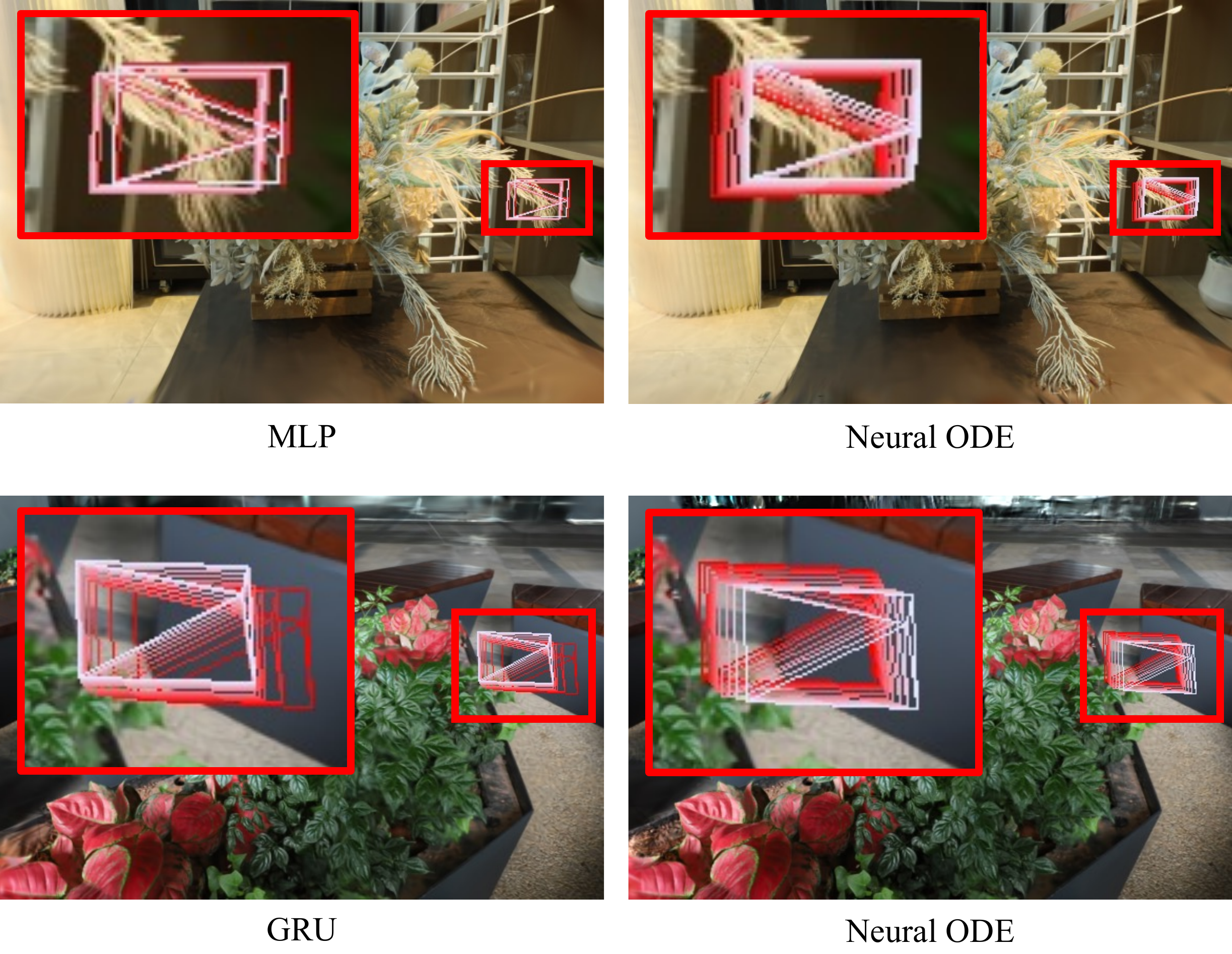}
	\caption{Comparison between the Neural ODE-based estimator and other estimators. Only the Neural ODE-based estimator exhibits a continuous camera trajectory.}
	\label{fig:3_neuralode_vs}
\end{figure}

\paragraph{Visualization for Other Estimators. }
To provide a qualitative evaluation of the camera motion estimators in \cref{tab:4_ablation_ode} of our main paper, we visualize the camera poses generated by the MLP-based and GRU-based estimators in \cref{fig:3_neuralode_vs}.

Since the MLP estimator does not account for the sequential nature of camera motion, it inherently lacks continuity, resulting in a trajectory that appears discontinuous and inconsistent. 
While the GRU estimator produces trajectories that seem more continuous, it separately implements GRU cells for forward and backward propagation, causing discontinuities between camera poses along the trajectory.

In contrast, our neural ODE-based estimator ensures a fully continuous camera motion over time, achieving both visually smooth trajectories and superior quantitative performance compared to the MLP and GRU estimators.

\begin{figure}[t]
	\centering
	\includegraphics[width=\linewidth]{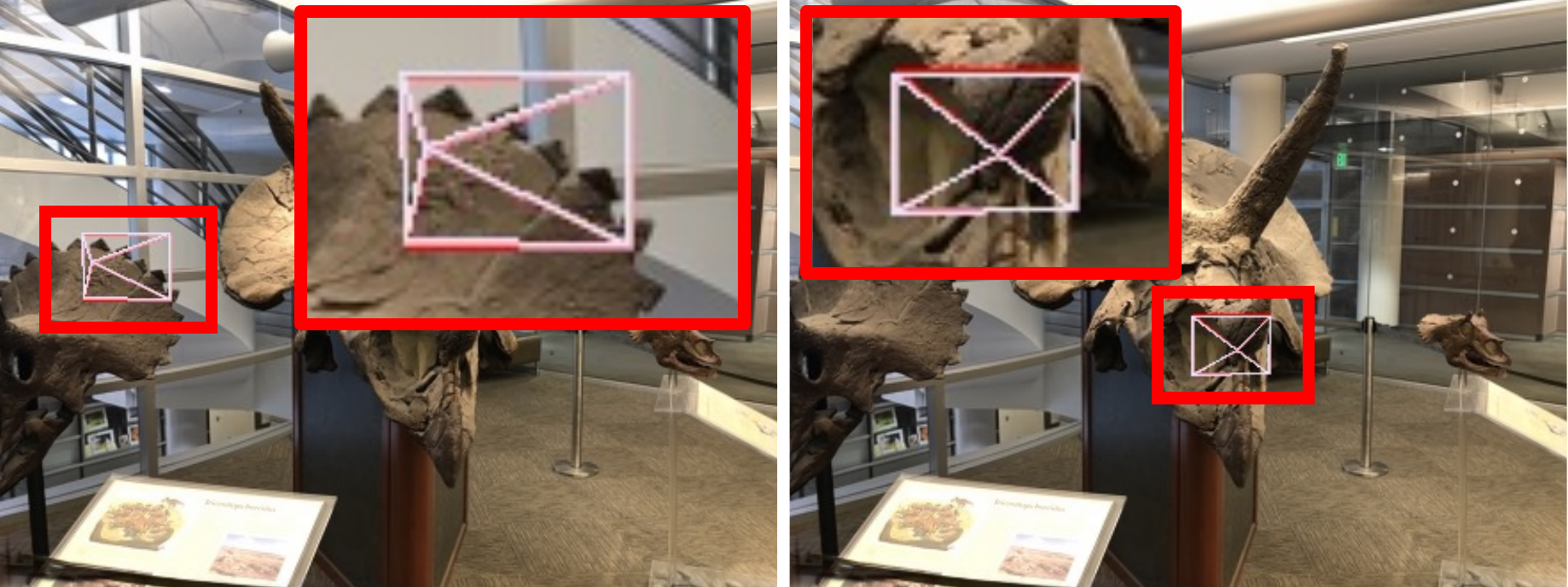}
	\caption{Visualization of camera motion trajectories for sharp images.}
	\label{fig:4_llff}
\end{figure}

\paragraph{Visualization for Sharp Images. }
We visualize the camera poses to examine how CoMoGaussian's camera trajectory modeling operates on the NeRF-LLFF dataset~\cite{mildenhall2019local,mildenhall2020nerf}, which consists of sharp images. As shown in \cref{fig:4_llff}, the camera motion trajectory for sharp images remains nearly stationary, demonstrating the generalization capability of our proposed method.

\begin{figure*}[t]
	\centering
	\includegraphics[width=0.9\linewidth]{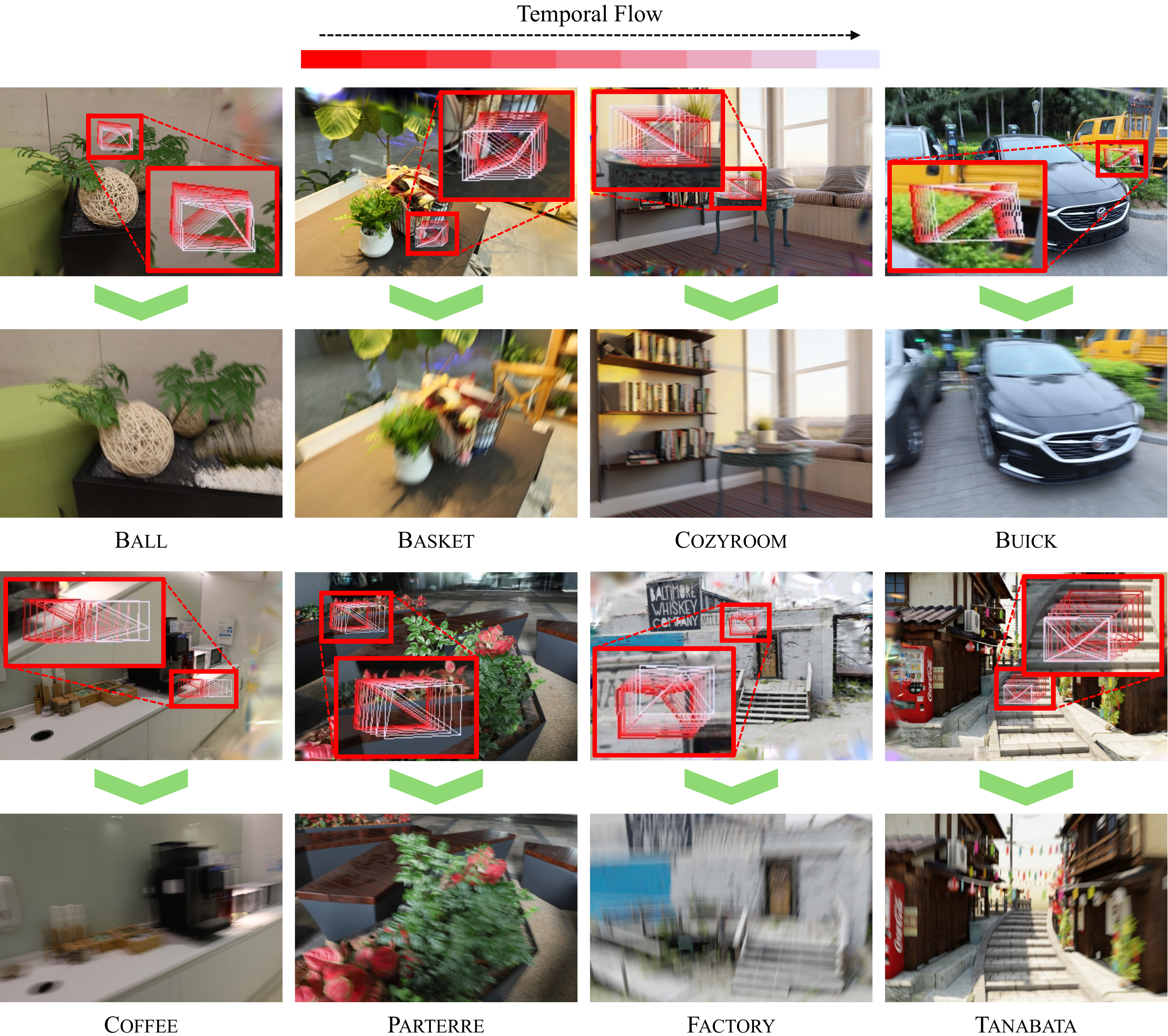}
	\caption{Camera motion trajectory predicted by CoMoGaussian for input images with significant blur.}
	\label{fig:kernel_visualize_extreme}
\end{figure*}

\begin{figure*}[t]
	\centering
	\includegraphics[width=0.9\linewidth]{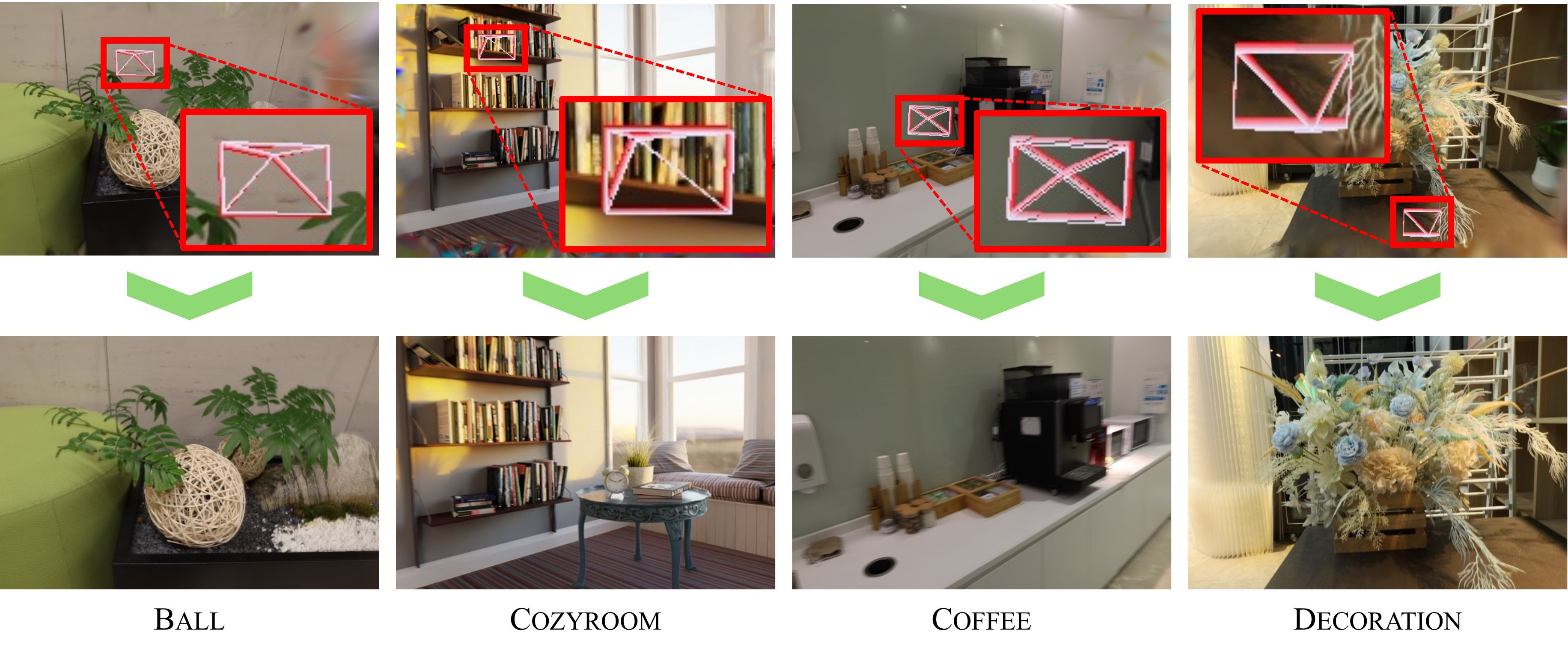}
	\caption{Camera motion trajectory predicted by CoMoGaussian for input images with moderate blur.}
	\label{fig:kernel_visualize_moderate}
\end{figure*}

\paragraph{Additional Visualization}
We visualize the camera motion trajectories for input blurry images predicted by CoMoGaussian in \cref{fig:kernel_visualize_extreme} and \cref{fig:kernel_visualize_moderate}. 
\cref{fig:kernel_visualize_extreme} illustrates camera motions for images with significant blur, where the predicted trajectories are continuous over time and align precisely with the input images. 
\cref{fig:kernel_visualize_moderate} depicts camera motions for images with relatively less blur, where the predicted trajectories show minimal movement, yet still match the input images accurately. 
These results demonstrate that our blurring kernel effectively models precise continuous camera motion.

\begin{table}[h]
	\caption{Comparison of training and rendering speeds across various 3DGS-based methods. * indicates that the speed is identical to that of the corresponding model.}
	\vspace{-3mm}
	\begin{center}
		\resizebox{\columnwidth}{!}{
			\centering
			\setlength{\tabcolsep}{7pt}
			\begin{tabular}{l||c|c}
				\toprule 
				Methods					& Training Time (hours)		& Rendering Speed  \\ \midrule
				BAD-Gaussians~\cite{zhao2024bad}						& 0.37			& *3DGS~\cite{kerbl20233d} \\
				Deblurring 3DGS~\cite{lee2024deblurring}				& 0.20			& *3DGS~\cite{kerbl20233d}\\
				BAGS~\cite{peng2024bags}										& 0.83			& *Mip-Splatting~\cite{yu2024mip}\\ \midrule
				\textbf{CoMoGaussian}												& 1.33			& *Mip-Splatting~\cite{yu2024mip}\\ \bottomrule
			\end{tabular}
		}
	\end{center}
	\label{tab:2_time}
	\vspace{-7mm}
\end{table}

\section{Training and Rendering Speed}
In this section, we compare the training time and rendering speed of our method with recent 3DGS-based approaches. 
As shown in \cref{tab:2_time}, CoMoGaussian requires a longer training time compared to other methods. 
However, given the quantitative results in \cref{tab:real,tab:synthetic,tab:exblurf_real}, as well as the qualitative comparisons in our \textit{supplementary videos}, our contributions remain significant. 
We believe that addressing the limitations discussed in the main paper will enable faster training in the future.

Additionally, since sharp rendering is performed solely using 3DGS~\cite{kerbl20233d} or Mip-Splatting~\cite{yu2024mip} without additional modules, our method achieves fast rendering speeds comparable to those of other approaches.

\section{Derivation of Rigid Body Motion~\cite{lynch2017modernrobotics}}
In this section, we explain the derivation process for \cref{eq:9_SO3} and \cref{eq:10_G_theta} from the main paper. This derivation aims to expand and simplify the process described in Modern Robotics~\cite{lynch2017modernrobotics} for better clarity and accessibility.

The components of a given screw axis include the unit rotation axis $\hat{\omega}\in\mathbb{R}^{3}$ and the translation component $v\in\mathbb{R}^{3}$. The unit rotation axis consists of the angular velocity $\omega$ and the rotation angle $\theta$:
\begin{equation}
	\hat{\omega}=\frac{\omega}{\theta},\quad where\quad||\hat{\omega}||=1.
\end{equation}
We combine the rotation axis $\hat{\omega}$ and the rotation angle $\theta$ to represent an element of the Lie Algebra, $\mathfrak{so}(3)$, which serves as the linear approximation of the rotation matrix. Before proceeding, $\hat{\omega}$ is converted into a $3\times 3$ skew-symmetric matrix $[\hat{\omega}]$ to compactly express the cross-product operation as a matrix multiplication:
\begin{equation}\label{eq:skew}
	[\hat{\omega}]=\left[
		\begin{array}{ccc}
			0 & -\hat{\omega}_{z} & \hat{\omega}_{y} \\
			\hat{\omega}_{z} & 0 & -\hat{\omega}_{x} \\
			-\hat{\omega}_{y} & \hat{\omega}_{x} & 0 \\
		\end{array}	
	\right] 
	\in \mathfrak{so}(3),~where~ [\hat{\omega}]^3 = -[\hat{\omega}].
\end{equation}
Using the skew-symmetric matrix $[\hat{\omega}]$ and the translation component $v$, the screw axis $[S]$ is expressed. By multiplying this screw axis with $\theta$, we incorporate the magnitude of the rotation and translation along the screw axis:
\begin{equation}
	[\mathcal{S}]\theta=\left[
		\begin{array}{cc}
			[\hat{\omega}]\theta & v\theta \\
			0 & 0 \\
		\end{array}
	\right]
	\in \mathfrak{se}(3),
\end{equation}
where $\mathfrak{se}(3)$ represents the Lie Algebra, which corresponds to the infinitesimal changes of the Lie Group \textit{SE}(3). To map this infinitesimal change to the \textit{SE}(3) transformation matrix $\mathbf{T}=e^{[\mathcal{S}]\theta}$, we use the Taylor expansion, following these steps:\\
{\small{\begin{align}
	e^{[\mathcal{S}]\theta} & = \sum_{n=0}^{\infty} [\mathcal{S}]^{n}\frac{\theta^{n}}{n!} \\
											  & = I + [\mathcal{S}]\theta + [\mathcal{S}]^2\frac{\theta^2}{2!} + \cdots \\
											  	& = \left[\begin{array}{cc}
											  		I + [\hat{\omega}]\theta + [\hat{\omega}]^{2}\frac{\theta^{2}}{2!}+\cdots & \left(I\theta + [\hat{\omega}]\frac{\theta^{2}}{2!}+\cdots\right)v \\
											  		0 & 1 \\
											  	\end{array}
											  	\right] \\
											  	& = \left[\begin{array}{cc}
											  		e^{[\hat{\omega}]\theta} & G(\theta)v \\
											  		0 & 1 \\
											  		\end{array}
											  	\right] \in \textit{SE}(3)\\
											  	& \because~[\mathcal{S}]^{n}=\left[
											  	\begin{array}{cc}
											  		[\hat{\omega}]^{n} & [\hat{\omega}]^{n-1}v \\
											  		0 & 0 \\
											  	\end{array}
											  	\right]
\end{align}}}
For the rotation matrix $e^{[\hat{\omega}]}\theta$, we simplify it using the Taylor expansion and \cref{eq:skew}, resulting in:
{\small{\begin{align}
	e^{[\hat{\omega}]\theta} & = I + [\hat{\omega}]\theta + [\hat{\omega}]^{2}\frac{\theta^{2}}{2!} + [\hat{\omega}]^{3}\frac{\theta^{3}}{3!} +  + [\hat{\omega}]^{4}\frac{\theta^{4}}{4!}\cdots \\
											  & = I + \left(\theta - \frac{\theta^{3}}{3!} + \cdots\right)[\hat{\omega}] + \left(\frac{\theta^{2}}{2!} - \frac{\theta^{4}}{4!} + \cdots\right)[\hat{\omega}]^{2} \\
											  & = I + \sin \theta[\hat{\omega}] + (1 - \cos \theta)[\hat{\omega}]^2 \in \textit{SO}(3)
\end{align}}}
The translational component $G(\theta)$ is also derived using the Taylor expansion and \cref{eq:skew}:
{\small{\begin{align}
	G(\theta) & = I\theta + [\hat{\omega}]\frac{\theta^{2}}{2!} + [\hat{\omega}]^{2}\frac{\theta^{3}}{3!} + [\hat{\omega}]^{3}\frac{\theta^{4}}{4!}+\cdots \\
	& = I\theta + \left(\frac{\theta^{2}}{2!} - \frac{\theta^{4}}{4!} + \cdots\right)[\hat{\omega}] + \left(\frac{\theta^{3}}{3!} - \frac{\theta^{5}}{5!} + \cdots\right)[\hat{\omega}]^{2} \\
	& = I\theta + (1 - \cos \theta)[\hat{\omega}] + (\theta - \sin \theta)[\hat{\omega}]^2
\end{align}}}
The term $G(\theta)$ physically represents the total translational motion caused by the rotational motion as the rigid body rotates by $\theta$. In other words, $G(\theta)$ indicates how rotational motion contributes to translational motion, which can also be expressed as an integral of the rotation motion:
\begin{align}
	G(\theta) & = \int_{0}^{\theta} e^{[\hat{\omega}]\theta}d\theta \\
	& = \int_{0}^{\theta} \left(I + \sin \theta[\hat{\omega}] + (1 - \cos \theta)[\hat{\omega}]^2\right)d\theta \\
	& = I\theta + (1 - \cos \theta)[\hat{\omega}] + (\theta - \sin \theta)[\hat{\omega}]^2
\end{align}
Through the above process, we derive \cref{eq:9_SO3} and \cref{eq:10_G_theta} in the main paper, improving readability of the paper and providing a clear foundation for understanding the mathematical framework.

\section{Per-Scene Quantitative Results}
We show the per-scene quantitative performance on Deblur-NeRF real-world, synthetic, and ExbluRF real-world dataset  in \cref{tab:real}, \cref{tab:synthetic}, and \cref{tab:exblurf_real}. CoMoGaussian demonstrates superior quantitative performance on most scenes in the real-world dataset.

\section{Additional Qualitative Results}
\begin{figure*}[t]
	\centering
	\includegraphics[width=\linewidth]{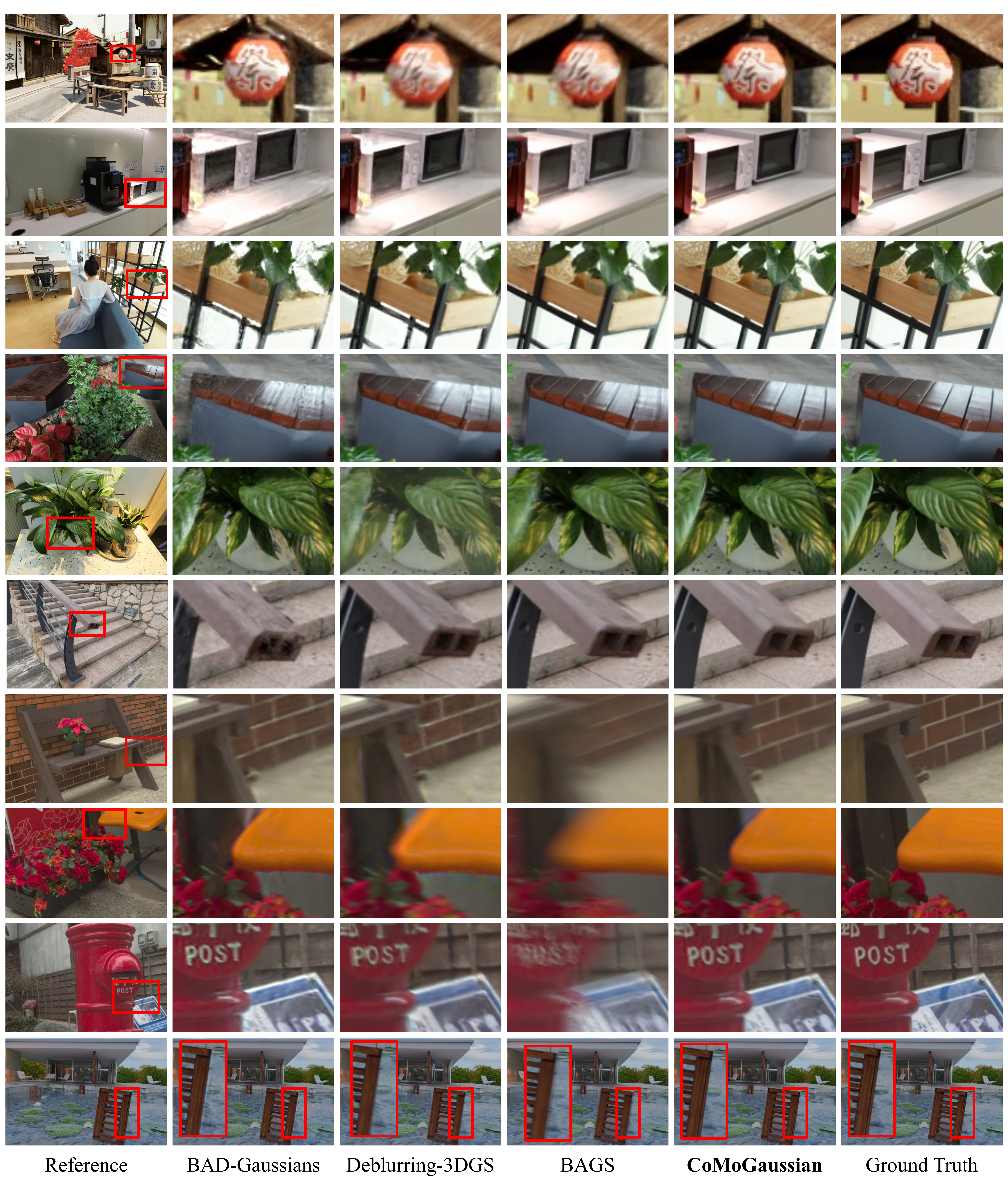}
	\caption{Additional Qualitative Comparison on the Synthetic and Real-World Scenes.}
	\label{fig:additional}
\end{figure*}

We provide additional visualization results in \cref{fig:additional}, which demonstrate that our CoMoGaussian outperforms not only in quantitative metrics but also in qualitative performance. For comparative videos, please refer to \textit{the supplementary materials}.

\begin{table*}[t]
	\caption{Per-Scene Quantitative Performance on Deblur-NeRF Real-World Scenes.}
	\vspace{-5mm}
	\begin{center}
		\resizebox{\linewidth}{!}{
			\centering
			\setlength{\tabcolsep}{1.5pt}
			\scriptsize
			\begin{tabular}{l||c|c|c|c|c|c|c|c|c|c|c|c|c|c|c}
				\toprule 

				\multirow{2}{*}{Real-World Scene\ } 			   & \multicolumn{3}{c|}{\textsc{Ball}}  	   & \multicolumn{3}{c|}{\textsc{Basket}}  	   & \multicolumn{3}{c|}{\textsc{Buick}}  				& \multicolumn{3}{c|}{\textsc{Coffee}} 		 & \multicolumn{3}{c}{\textsc{Decoration}}  \\ \cmidrule{2-16}
				&\ PSNR$\uparrow$ \     &\ SSIM$\uparrow$ \     &\ LPIPS$\downarrow$ \  &\ PSNR$\uparrow$ \     &\ SSIM$\uparrow$ \     &\ LPIPS$\downarrow$ \ 		&\ PSNR$\uparrow$ \     &\ SSIM$\uparrow$ \     &\ LPIPS$\downarrow$ \ 		&\ PSNR$\uparrow$ \     &\ SSIM$\uparrow$ \       &\ LPIPS$\downarrow$ \ 	&\ PSNR$\uparrow$ \     &\ SSIM$\uparrow$ \       &\ LPIPS$\downarrow$ \   	\\ \midrule \midrule
				Naive NeRF~\cite{mildenhall2020nerf}                            				 & 24.08    & 0.6237   & 0.3992 		& 23.72       & 0.7086      & 0.3223      & 21.59    & 0.6325      & 0.3502    		& 26.48    & 0.8064      & 0.2896    		& 22.39    & 0.6609     & 0.3633    \\
				Mip-Splatting~\cite{yu2024mip}                            				 & 23.22 & 0.6190 & 0.3400 & 23.24 & 0.6880 & 0.2880 & 21.46 & 0.6590 & 0.2660 & 24.73 & 0.7490 & 0.2880 & 20.55 & 0.6410 & 0.2990  \\ \midrule
				Deblur-NeRF~\cite{ma2022deblurnerf}                         				& 27.36    & 0.7656   & 0.2230 		 & 27.67       & 0.8449      & 0.1481        & 24.77     & 0.7700      & 0.1752    		& 30.93     & 0.8981      & 0.1244    		 & 24.19    & 0.7707     & 0.1862    \\
				DP-NeRF~\cite{lee2023dp}                        						  & 27.20    & 0.7652   & 0.2088 		& 27.74   	 & 0.8455      & 0.1294    		& 25.70    & 0.7922      & 0.1405    	& 31.19    & 0.9049      & 0.1002    		& 24.31    & 0.7811     & 0.1639   \\ \midrule
				BAD-Gaussians~\cite{zhao2024bad}                            				 & 22.28 & 0.6032 & 0.2054 & 22.02 & 0.7004 & 0.1197 & 19.95 & 0.6127 & 0.1103 & 25.58 & 0.7965 & 0.0932 & 21.11 & 0.6651 & 0.1185    \\ 
				Deblurring 3DGS~\cite{lee2024deblurring}												& \cellcolor{second!35}28.27 & \cellcolor{second!35}0.8233 & \cellcolor{second!35}0.1413 & 28.42 & 0.8713 & 0.1155		& 25.95		& 0.8367		& 0.0954		& \cellcolor{second!35}32.84		& 0.9312 & 	\cellcolor{second!35}0.0676	& 25.87 & 0.8540 &  0.0933 \\
				BAGS~\cite{peng2024bags}                            				 & 27.68 & 0.7990 & 0.1500 & \cellcolor{second!35}29.54 & \cellcolor{second!35}0.9000 & \cellcolor{best!25}0.0680 & \cellcolor{second!35}26.18 & \cellcolor{second!35}0.8440 & \cellcolor{second!35}0.0880 & 31.59 & 0.9080 & 0.0960 & \cellcolor{second!35}26.09 & \cellcolor{second!35}0.8580 & \cellcolor{best!25}0.0830    \\ \midrule
				\textbf{CoMoGaussian}                            				 & \cellcolor{best!25}29.60 & \cellcolor{best!25}0.8422 & \cellcolor{best!25}0.1115 & \cellcolor{best!25}30.78 & \cellcolor{best!25}0.9041 & \cellcolor{second!35}0.0761 & \cellcolor{best!25}27.23 & \cellcolor{best!25}0.8502 & \cellcolor{best!25}0.0742 & \cellcolor{best!25}33.04 & \cellcolor{second!35}0.9247 & \cellcolor{best!25}0.0578 & \cellcolor{best!25}26.44 & \cellcolor{best!25}0.8601 & \cellcolor{second!35}0.0891  \\ \midrule \midrule
				
				\multirow{2}{*}{Real-World Scene\ } 			   & \multicolumn{3}{c|}{\textsc{Girl}}  	   & \multicolumn{3}{c|}{\textsc{Heron}}  	   & \multicolumn{3}{c|}{\textsc{Parterre}}  				& \multicolumn{3}{c|}{\textsc{Puppet}} 		 & \multicolumn{3}{c}{\textsc{Stair}}  \\ \cmidrule{2-16}
				&\ PSNR$\uparrow$ \     &\ SSIM$\uparrow$ \     &\ LPIPS$\downarrow$ \  &\ PSNR$\uparrow$ \     &\ SSIM$\uparrow$ \     &\ LPIPS$\downarrow$ \ 		&\ PSNR$\uparrow$ \     &\ SSIM$\uparrow$ \     &\ LPIPS$\downarrow$ \ 		&\ PSNR$\uparrow$ \     &\ SSIM$\uparrow$ \       &\ LPIPS$\downarrow$ \ 	&\ PSNR$\uparrow$ \     &\ SSIM$\uparrow$ \       &\ LPIPS$\downarrow$ \   	\\ \midrule \midrule
				Naive NeRF~\cite{mildenhall2020nerf}                            				 & 20.07    & 0.7075   & 0.3196 		& 20.50       & 0.5217      & 0.4129      & 23.14    & 0.6201      & 0.4046    		& 22.09    & 0.6093      & 0.3389    		& 22.87    & 0.4561     & 0.4868    \\ 
				Mip-Splatting~\cite{yu2024mip}                            				 & 19.87 & 0.7140 & 0.2780 & 19.43 & 0.5050 & 0.3320 & 22.28 & 0.5900 & 0.3210 & 22.05 & 0.6310 & 0.2670 & 21.91 & 0.4740 & 0.3870   \\ \midrule
				Deblur-NeRF~\cite{ma2022deblurnerf}                         				& 22.27    & 0.7976   & 0.1687 		 & 22.63       & 0.6874      & 0.2099        & 25.82     & 0.7597      & 0.2161    		& 25.24     & 0.7510      & 0.1577    		 & 25.39    & 0.6296     & 0.2102    \\
				DP-NeRF~\cite{lee2023dp}                        						  & 23.33    & 0.8139   & 0.1498 		& 22.88   	 & 0.6930      & 0.1914    		& 25.86    & 0.7665      & 0.1900    	& 25.25    & 0.7536      & 0.1505    		& 25.59    & 0.6349      & 0.1772   \\ \midrule 
				BAD-Gaussians~\cite{zhao2024bad}                            				 	& 19.16 & 0.7037 & 0.1178 & 19.47 & 0.5264 & 0.1747 & 21.71 & 0.6154 & 0.1165 & 21.74 & 0.6506 & 0.1166 & 23.86 & 0.5969 & 0.0892  \\ 
				Deblurring 3DGS~\cite{lee2024deblurring}							& 23.26 & 0.8390 & 	0.1011	& \cellcolor{second!35}23.14 & \cellcolor{best!25}0.7438 & 0.1543	 		& \cellcolor{best!25}26.17 & \cellcolor{second!35}0.8144	& 0.1206		& 25.67 & \cellcolor{second!35}0.8051	& 0.0941		& 26.46 & 0.7050 & 0.1123 \\
				BAGS~\cite{peng2024bags}                            				 		& \cellcolor{second!35}25.45 & \cellcolor{second!35}0.8690 & \cellcolor{second!35}0.0790 & 22.04 & 0.7150 & \cellcolor{best!25}0.1260 & 25.92 & \cellcolor{best!25}0.8190 & \cellcolor{second!35}0.0920 & \cellcolor{second!35}25.81 & 0.8040 & \cellcolor{second!35}0.0940 & \cellcolor{second!35}26.69 & \cellcolor{second!35}0.7210 & \cellcolor{second!35}0.0800 \\ \midrule
				\textbf{CoMoGaussian}                            				 	& \cellcolor{best!25}27.08 & \cellcolor{best!25}0.8884 & \cellcolor{best!25}0.0733 & \cellcolor{best!25}23.18 & \cellcolor{second!35}0.7350 & \cellcolor{second!35}0.1326 & \cellcolor{second!35}26.11 & 0.8131 & \cellcolor{best!25}0.0872 & \cellcolor{best!25}27.09 & \cellcolor{best!25}0.8337 & \cellcolor{best!25}0.0658 & \cellcolor{best!25}27.92 & \cellcolor{best!25}0.7792 & \cellcolor{best!25}0.0541   \\ \bottomrule
			\end{tabular}
		}
	\end{center}
	\label{tab:real}
	\vspace{-4mm}
\end{table*}

\begin{table*}[t]
	\caption{Per-Scene Quantitative Performance on Deblur-NeRF Synthetic Scenes.}
	\vspace{-5mm}
	\begin{center}
		\resizebox{\linewidth}{!}{
			\centering
			\setlength{\tabcolsep}{1.5pt}
			\scriptsize
			\begin{tabular}{l||c|c|c|c|c|c|c|c|c|c|c|c|c|c|c}
				\toprule 
				
				\multirow{2}{*}{Synthetic Scene} 			   & \multicolumn{3}{c|}{\textsc{Factory}}  	   & \multicolumn{3}{c|}{\textsc{CozyRoom}}  	   & \multicolumn{3}{c|}{\textsc{Pool}}  				& \multicolumn{3}{c|}{\textsc{Tanabata}} 		 & \multicolumn{3}{c}{\textsc{Trolley}}  \\ \cmidrule{2-16}
				&\ PSNR$\uparrow$ \     &\ SSIM$\uparrow$ \     &\ LPIPS$\downarrow$ \  &\ PSNR$\uparrow$ \     &\ SSIM$\uparrow$ \     &\ LPIPS$\downarrow$ \ 		&\ PSNR$\uparrow$ \     &\ SSIM$\uparrow$ \     &\ LPIPS$\downarrow$ \ 		&\ PSNR$\uparrow$ \     &\ SSIM$\uparrow$ \       &\ LPIPS$\downarrow$ \ 	&\ PSNR$\uparrow$ \     &\ SSIM$\uparrow$ \       &\ LPIPS$\downarrow$ \   	\\ \midrule \midrule
				Naive NeRF~\cite{mildenhall2020nerf}                            				 &19.32 & 0.4563 & 0.5304 & 25.66 & 0.7941 & 0.2288 & 30.45 & 0.8354 & 0.1932 & 22.22 & 0.6807 & 0.3653 & 21.25 & 0.6370 & 0.3633 \\ 
				Mip-Splatting~\cite{yu2024mip}                            				 & 18.21 & 0.4234 & 0.4769 & 25.25 & 0.7968 & 0.1646 & 30.57 & 0.8483 & 0.1456 & 21.54 & 0.6754 & 0.3075 & 20.82 & 0.6388 & 0.3165  \\ \midrule
				Deblur-NeRF~\cite{ma2022deblurnerf}                         				& 25.60    & 0.7750   & 0.2687 		 & 32.08       & 0.9261      & 0.0477        & 31.61     & 0.8682      & 0.1246    		& 27.11     & 0.8640      & 0.1228    		 & 27.45    & 0.8632     & 0.1363    \\
				DP-NeRF~\cite{lee2023dp}                        				 	 & \cellcolor{second!35}25.91    & \cellcolor{second!35}0.7787   & 0.2494 		& \cellcolor{second!35}32.65   	 & 0.9317      & 0.0355    		& \cellcolor{second!35}31.96    & 0.8768      & 0.0908    	& \cellcolor{second!35}27.61    & 0.8748      & 0.1033    		& 28.03    & 0.8752      & 0.1129   \\ \midrule
				BAD-Gaussians~\cite{zhao2024bad}										 	 & 17.86 & 0.3892 & \cellcolor{second!35}0.1440 & 23.50 & 0.7396 & 0.0616 & 26.90 & 0.7296 & 0.1127 & 20.54 & 0.6379 & 0.0860 & 21.26 & 0.6921 & \cellcolor{second!35}0.0963    \\ 
				Deblurring 3DGS~\cite{lee2024deblurring}												& 24.01    & 0.7333   & 0.2326 			& 31.45       & 0.9222      & 0.0367      & 31.87    & \cellcolor{second!35}0.8829      & \cellcolor{second!35}0.0751    		& 27.01    & \cellcolor{second!35}0.8807      & \cellcolor{second!35}0.0785    		& \cellcolor{second!35}26.88    & \cellcolor{second!35}0.8710     & 0.1028    \\
				BAGS~\cite{peng2024bags}												& 22.35 & 0.6639 & 0.2277 & 32.21 & \cellcolor{second!35}0.9359 & \cellcolor{second!35}0.0245 & 28.72 & 0.8404 & 0.0804 & 26.79 & 0.8735 & 0.1099 & 26.61 & 0.8627 & 0.1156  \\ \midrule \midrule
				\textbf{CoMoGaussian}											 & \cellcolor{best!25}29.32 & \cellcolor{best!25}0.8971 & \cellcolor{best!25}0.0563 & \cellcolor{best!25}33.34 & \cellcolor{best!25}0.9427 & \cellcolor{best!25}0.0239 & \cellcolor{best!25}32.45 & \cellcolor{best!25}0.8924 & \cellcolor{best!25}0.0705 & \cellcolor{best!25}29.53 & \cellcolor{best!25}0.9273 & \cellcolor{best!25}0.0408 & \cellcolor{best!25}30.45 & \cellcolor{best!25}0.9240 & \cellcolor{best!25}0.0546 \\  \bottomrule
			\end{tabular}
		}
		\label{tab:synthetic}
		\vspace{-4mm}
	\end{center}
\end{table*}

\begin{table*}[t]
	\caption{Per-Scene Quantitative Performance on the ExbluRF Real-World Scenes.}
	\vspace{-5mm}
	\begin{center}
		\resizebox{0.9\linewidth}{!}{
			\centering
			\setlength{\tabcolsep}{1.5pt}
			\scriptsize
			\begin{tabular}{l||c|c|c|c|c|c|c|c|c|c|c|c}
				\toprule 
				
				\multirow{2}{*}{ExbluRF\ } 			   & \multicolumn{3}{c|}{\textsc{Bench}}  	   & \multicolumn{3}{c|}{\textsc{Camellia}}  	   & \multicolumn{3}{c|}{\textsc{Dragon}}  				& \multicolumn{3}{c}{\textsc{Jars}}   \\ \cmidrule{2-13}
				&\ PSNR$\uparrow$ \     &\ SSIM$\uparrow$ \     &\ LPIPS$\downarrow$ \  &\ PSNR$\uparrow$ \     &\ SSIM$\uparrow$ \     &\ LPIPS$\downarrow$ \ 		&\ PSNR$\uparrow$ \     &\ SSIM$\uparrow$ \     &\ LPIPS$\downarrow$ \ 		&\ PSNR$\uparrow$ \     &\ SSIM$\uparrow$ \       &\ LPIPS$\downarrow$ \ 	\\ \midrule \midrule
				Mip-Splatting~\cite{yu2024mip}                            				 & 24.58 & 0.5671 & 0.6190 & 23.28 & 0.5151 & 0.5886 & 28.65 & 0.5403 & 0.7002 & 24.08 & 0.5335 & 0.6094 \\ \midrule
				ExbluRF~\cite{lee2023exblurf}                            				 & 24.75 & 0.5783 & 0.3003 & 23.14 & 0.4925 & 0.3630 & 24.26 & 0.4042 & 0.5641 & 22.21 & 0.4591 & 0.4213 \\ \midrule
				BAD-Gaussians~\cite{zhao2024bad}                            	  & 28.27 & 0.7125 & \cellcolor{second!35}0.2266 & 23.39 & 0.5102 & \cellcolor{second!35}0.3034 & 30.24 & 0.6383 & \cellcolor{best!25}0.4374 & \cellcolor{second!35}28.41 & \cellcolor{second!35}0.7041 & \cellcolor{best!25}0.3347 \\ 
				Deblurring 3DGS~\cite{lee2024deblurring}						& \cellcolor{second!35}30.44 & \cellcolor{second!35}0.7708 & 0.2587 & \cellcolor{second!35}26.26 & \cellcolor{second!35}0.6401 & 0.3964 & \cellcolor{second!35}30.87 & \cellcolor{second!35}0.6643 & 0.5561 & 27.56 & 0.6559 & 0.4431 \\
				BAGS~\cite{peng2024bags}                            				 & 25.40 & 0.6142 & 0.4962 & 23.29 & 0.5177 & 0.5450 & 29.06 & 0.5577 & 0.6738 & 24.00 & 0.5402 & 0.5534 \\ \midrule
				\textbf{CoMoGaussian}                            				 					& \cellcolor{best!25}31.82 & \cellcolor{best!25}0.8011 & \cellcolor{best!25}0.2170 & \cellcolor{best!25}28.53 & \cellcolor{best!25}0.7004 & \cellcolor{best!25}0.2846 & \cellcolor{best!25}31.95 & \cellcolor{best!25}0.7166 & \cellcolor{second!35}0.4476 & \cellcolor{best!25}29.63 & \cellcolor{best!25}0.7351 & \cellcolor{second!35}0.3474 \\ \midrule \midrule
				
				\multirow{2}{*}{ExbluRF\ } 			   & \multicolumn{3}{c|}{\textsc{Jars2}}  	   & \multicolumn{3}{c|}{\textsc{Postbox}}  	   & \multicolumn{3}{c|}{\textsc{Stone Lantern}}  				& \multicolumn{3}{c}{\textsc{Sunflowers}}   \\ \cmidrule{2-13}
				&\ PSNR$\uparrow$ \     &\ SSIM$\uparrow$ \     &\ LPIPS$\downarrow$ \  &\ PSNR$\uparrow$ \     &\ SSIM$\uparrow$ \     &\ LPIPS$\downarrow$ \ 		&\ PSNR$\uparrow$ \     &\ SSIM$\uparrow$ \     &\ LPIPS$\downarrow$ \ 		&\ PSNR$\uparrow$ \     &\ SSIM$\uparrow$ \       &\ LPIPS$\downarrow$ \ 	\\ \midrule \midrule
				Mip-Splatting~\cite{yu2024mip}                            				 & 22.10 & 0.5682 & 0.5817 & 23.19 & 0.5277 & 0.5710 & 22.64 & 0.6004 & 0.6308 & 25.78 & 0.6775 & 0.5020 \\ \midrule
				ExbluRF~\cite{lee2023exblurf}                            				 &21.99 & 0.5736 & 0.3519 & 23.34 & 0.5287 & 0.2978 & \cellcolor{second!35}26.18 & \cellcolor{second!35}0.6832 & 0.4236 & 25.25 & 0.6765 & 0.3223	\\ \midrule
				BAD-Gaussians~\cite{zhao2024bad}                            	 & 26.27 & 0.6914 & \cellcolor{second!35}0.3326 & \cellcolor{second!35}25.01 & \cellcolor{second!35}0.6264 & \cellcolor{second!35}0.2760 & 25.19 & 0.6724 & \cellcolor{second!35}0.3794 & 27.82 & 0.7443 & \cellcolor{second!35}0.2865 \\ 
				Deblurring 3DGS~\cite{lee2024deblurring}						& \cellcolor{second!35}26.76 & \cellcolor{second!35}0.7100 & 0.3942 & 23.89 & 0.5563 & 0.3492 & 23.32 & 0.6430 & 0.4687 & \cellcolor{second!35}29.75 & \cellcolor{second!35}0.7955 & 0.3248 \\
				BAGS~\cite{peng2024bags}                            				 & 22.20 & 0.5658 & 0.5268 & 24.76 & 0.5891 & 0.4205 & 22.72 & 0.5979 & 0.5556 & 26.14 & 0.6920 & 0.4508 \\ \midrule
				\textbf{CoMoGaussian}                            				 			& \cellcolor{best!25}29.72 & \cellcolor{best!25}0.7699 & \cellcolor{best!25}0.3283 & \cellcolor{best!25}29.99 & \cellcolor{best!25}0.7631 & \cellcolor{best!25}0.2479 & \cellcolor{best!25}28.66 & \cellcolor{best!25}0.7549 & \cellcolor{best!25}0.3402 & \cellcolor{best!25}30.90 & \cellcolor{best!25}0.8062 & \cellcolor{best!25}0.2622 \\ \bottomrule
			\end{tabular}
		}
	\end{center}
	\label{tab:exblurf_real}
\end{table*}

\end{document}